\documentclass{article}

\usepackage{PRIMEarxiv}

\usepackage[utf8]{inputenc} % allow utf-8 input
\usepackage[T1]{fontenc}    % use 8-bit T1 fonts
\usepackage{hyperref}       % hyperlinks
\usepackage{url}            % simple URL typesetting
\usepackage{booktabs}       % professional-quality tables
\usepackage{amsfonts}       % blackboard math symbols
\usepackage{nicefrac}       % compact symbols for 1/2, etc.
\usepackage{microtype}      % microtypography
\usepackage{lipsum}
\usepackage{fancyhdr}       % header
\usepackage{graphicx}       % graphics
\graphicspath{{media/}}     % organize your images and other figures under media/ folder
\usepackage{enumitem}
\usepackage{amsmath} % 加载 amsmath 宏包以支持更多数学环境

\usepackage{caption}
\usepackage{graphicx}
\usepackage{float} 
\usepackage{subcaption}
\usepackage{multirow}
\usepackage{cleveref}

\title{YOLOv11-RGBT: Towards a Comprehensive Single-Stage Multispectral Object Detection Framework
}

\author{
  Dahang Wan, Rongsheng Lu*, Yang Fang,  Xianli Lang,  Shuangbao Shu, Jingjing Chen, 
   Siyuan Shen,  Ting Xu  \\
  School of Instrument Science and Opto-electronics Engineering, \\
  Anhui Province Key Laboratory of Measuring Theory and Precision Instrument \\
  Hefei University of Technology \\
  Hefei 230009, China\\
  \texttt{\{wandahang, fangyoung\}@foxmail.com} \\
  \texttt{\{rslu, langxl, shu, jingjingchen\}@hfut.edu.cn} \\
  \texttt{\{shensiyuan, xuting\}@mail.hfut.edu.cn} \\
   \And
  Zecong Ye \\
  School of Information Engineering \\
  Engineering University of PAP \\
  Xi’an 710086, China\\
  \texttt{yzc6666@yeah.net} \\
}

\begin{document}
\maketitle

\begin{abstract}
% \lipsum[1]
Multispectral object detection, which integrates information from multiple bands, can enhance detection accuracy and environmental adaptability, holding great application potential across various fields. Although existing methods have made progress in cross-modal interaction, low-light conditions, and model lightweight, there are still challenges like the lack of a unified single-stage framework, difficulty in balancing performance and fusion strategy, and unreasonable modality weight allocation. To address these, based on the YOLOv11 framework, we present YOLOv11-RGBT, a new comprehensive multimodal object detection framework. We designed six multispectral fusion modes and successfully applied them to models from YOLOv3 to YOLOv12 and RT-DETR. After reevaluating the importance of the two modalities, we proposed a P3 mid-fusion strategy and multispectral controllable fine-tuning (MCF) strategy for multispectral models. These improvements optimize feature fusion, reduce redundancy and mismatches, and boost overall model performance. Experiments show our framework excels on three  major open-source multispectral object detection datasets, like LLVIP and FLIR. Particularly, the multispectral controllable fine-tuning strategy significantly enhanced model adaptability and robustness. On the FLIR dataset, it consistently improved YOLOv11 models' mAP by 3.41\%-5.65\%, reaching a maximum of 47.61\%, verifying the framework and strategies' effectiveness. The code is available at: \href{https://github.com/wandahangFY/YOLOv11-RGBT}{https://github.com/wandahangFY/YOLOv11-RGBT}.

\end{abstract}

% keywords can be removed
\renewcommand{\thefootnote}{}% 取消脚注标号
\footnote{*Corresponding Author}

% \mainmatter
\renewcommand{\thefootnote}{\arabic{footnote}}% 恢复默认的脚注标号格式

\section{Introduction}
Object detection, a key computer vision task, aims to identify and locate specific objects in images or videos \cite{wan_yolo-mif_2024}. Deep learning, especially CNN-based methods, has significantly advanced this field. However, traditional visible-light detection algorithms, which rely on RGB images, struggle in complex conditions like low light, bad weather, or camouflaged targets \cite{liu_coconet_2024}. They also can't capture multi-dimensional object features, limiting detection robustness and accuracy \cite{liu_multi-interactive_2023,liu_target-aware_2022}.

Multispectral imaging, capturing electromagnetic spectra beyond visible light (e.g., infrared, near-infrared, short-wave infrared), offers a solution \cite{hwang_multispectral_2015}. It provides richer object features, such as thermal radiation, vegetation health, and camouflage-penetration ability. These additional spectral details enhance detection performance, particularly in complex environments, driving the development of multispectral object detection algorithms that leverage these images to improve accuracy and robustness.

Early multispectral object detection methods used traditional RGB models like YOLO \cite{redmon_you_2016,redmon_yolo9000_2016,redmon_yolov3_2018,bochkovskiy_yolov4_2020,noauthor_ultralyticsyolov5_2022, li_yolov6_2022,wang_yolov7_2023,jocher_yolo_2023, wang_yolov9_2024, wang_yolov10_2024, jocher_yolo_2023,khanam_yolov11_2024,tian_yolov12_2025}, SSD \cite{leibe_ssd_2016,fu_dssd_nodate}, and R-CNN \cite{girshick_rich_2014,girshick_fast_2015,ren_faster_2017,cai_cascade_2018} directly on multispectral images. But their poor performance on multispectral data stemmed from underutilizing complementary information across spectral modalities. For instance, significant redundancy between RGB and infrared images led to information waste and insufficient performance when using traditional models. Consequently, researchers started exploring multispectral feature fusion methods.

Multispectral object detection feature fusion strategies are categorized into early, mid-level, and late decision-level fusion based on their processing stage[20]. Early fusion integrates multispectral information during data collection or initial feature extraction to enrich input features. Mid-level fusion occurs during backbone feature extraction, enhancing network expressiveness through intermodal feature interaction. Late decision-level fusion combines detection results from different modalities in the final detection stage to boost overall performance. These fusion methods mark a shift from simple multi-modal stacking to more efficient feature integration and information complementarity, laying the foundation for improved multispectral object detection.

Early fusion techniques comprise conventional image fusion methods \cite{li_pixel-level_2017} such as GRW (gradient-based region weighting) and GFF (gradient field fusion), as well as advanced deep learning-based approaches. For example, MDCNN \cite{dong_mdcnn_2021} improves image quality in multi-scale feature extraction and fusion, CrossFuse\cite{li_crossfuse_2024} enhances data robustness and generalization with Top-k visual alignment and self-supervised learning, and DIVFusion \cite{tang_divfusion_2023} optimizes infrared and visible image fusion using SIDNet and TCEFNet in an unsupervised manner. Despite their excellent performance, these deep-learning-based image fusion technologies are often computationally complex, time-consuming, and lack embeddability, making them more suitable for offline training. In multispectral object detection practice, there is an increasing trend towards mid-level fusion strategies. Studies \cite{li_illumination-aware_2019, liu_multispectral_2016} using Faster R-CNN as a baseline have revealed significant complementarity between visible and infrared light in pedestrian detection tasks. Researchers have designed various fusion methods, with Halfway Fusion standing out by effectively improving detection performance through fusion in the middle stage of feature extraction and being adopted in subsequent studies. However, due to the slow speed and high deployment costs of two-stage models, subsequent research has shifted more towards improved YOLO-based models. These improved models have further enhanced the efficiency and performance of multispectral object detection by optimizing architecture and fusion strategies. Early mid-level feature fusion methods \cite{sharma_yolors_2021} mainly used feature concatenation or addition, but these approaches suffered from feature misalignment and poor fusion performance. To address these issues, researchers introduced various cross-attention mechanisms. For instance, Cross-Modality Fusion Transformer (CFT) \cite{qingyun_cross-modality_2022} first applied Transformer to multispectral object detection, improving multispectral object detection performance of YOLOv5 and YOLOv3 by fusing visible and infrared features at each layer of the backbone network. Nevertheless, the huge number of parameters in CFT limits its efficiency in practical applications. To reduce model complexity, researchers have begun exploring more lightweight fusion methods \cite{sharma_yolors_2021,fang_cross-modality_2021}. For example, ICAFusion \cite{shen_icafusion_2024} proposed a dual cross-attention feature fusion method that maintains high detection performance with fewer parameters through an iterative interaction mechanism and a cross-modal feature enhancement module.

Subsequent research has delved into multifaceted aspects of multispectral object detection, including multispectral multi-scale feature fusion \cite{li_multiscale_2024}, modality imbalance \cite{zhou_improving_2020}, and low-light adaptation \cite{tang_piafusion_2022,zhang_illumination-guided_2023,tang_divfusion_2023}. By integrating Transformer's self-attention or conventional spatial attention mechanisms like CBAM\cite{woo_cbam_2018} and MLCA \cite{wan_mixed_2023}, researchers have effectively harnessed complementary information from visible and infrared images. This has led to superior performance on datasets like FLIR \cite{zhang_multispectral_2020}, M3FD \cite{liu_target-aware_2022}, and VEDAI \cite{razakarivony_vehicle_2016}, and robustness in complex conditions. However, in mid-level fusion studies \cite{qingyun_cross-modality_2022,noauthor_adopting_nodate,tang_piafusion_2022,zhou_improving_2020,yan_cross-modality_2023}, modalities are often treated as equally important, which is limiting. In reality, one modality usually has an edge in multispectral detection tasks. For instance, visible light outperforms infrared in the VEDAI dataset, while infrared is better for pedestrian detection in datasets like LLVIP \cite{jia_llvip_2021} and KAIST \cite{choi_kaist_2018}. This highlights the need for differentiated modality treatment and fusion strategy refinement in specific scenarios. Despite notable progress in multispectral object detection, particularly in cross-modal interaction, low-light conditions, and model lightweightness, several challenges persist:

\begin{enumerate}[label=(\arabic*)]
    \item  \textbf{Lack of Unified Framework:} Current methods are mostly model-specific or scene-specific, lacking a versatile single-stage multispectral detection framework. This limits algorithm generalizability and scalability across diverse applications.
    \item \textbf{Unreasonable Modality Weighting:} Most networks treat modalities as equally important. Yet, in practice, one modality often surpasses the other. Uniform feature fusion may degrade model performance, even below single-modality detection levels.
    \item \textbf{Balancing Model Performance and Fusion Strategy:} Selecting optimal fusion strategies across different stages remains challenging. Existing methods often fail to balance model performance and fusion effectively, compromising detection accuracy and efficiency.
    
\end{enumerate}

To address these challenges, this paper introduces YOLOv11-RGBT, a multimodal detection framework based on YOLOv11. It aims to balance detection accuracy, speed, and model parameters while maximizing feature utilization. The key contributions are:

\begin{enumerate}[label=(\arabic*)]
    \item \textbf{YOLOv11-RGBT:} A unified multispectral detection framework YOLOv11-RGBT supporting various tasks like detection, image classification, instance segmentation, and keypoint detection.
    \item \textbf{Rethinking multispectral feature mid-fusion  strategies:} Experiments show that mid-level fusion is suitable for single-stage detection. The proposed P3 mid-level fusion strategy achieves better results with fewer parameters by fusing at the right position once instead of multiple times.
    \item  \textbf{Multispectral controllable fine-tuning (MCF):} A controllable fine-tuning strategy for multispectral models inspired by ControlNet. It freezes pre-trained single-modal weights and introduces the other modality through fine-tuning to enhance detection stability.
    \item  \textbf{Six multispectral fusion modes:} Six designed single-stage multispectral fusion modes applied to multiple models, including YOLOv3-YOLOv12, PP-YOLOE, and RT-DETR, enabling multispectral task implementation across various single-stage networks.
\end{enumerate}

The paper is structured as follows: Section 2 reviews related work on multispectral object detection. Section 3 details the YOLOv11-RGBT framework and model components. Section 4 presents experimental results on three datasets. Section 5 discusses the experiments, and Section 6 concludes the study and outlines future work.

\section{Related Work}
\label{sec:headings}

\subsection{General object detection algorithms for multispectral detection}
Object detection models are crucial in multispectral detection, enabling automatic object identification and localization in multispectral images. In recent years, deep learning, particularly CNN-based models, has significantly improved detection efficiency and accuracy through specialized network structures and loss functions. These models can be divided into single-stage models (e.g., YOLO \cite{redmon_you_2016,redmon_yolo9000_2016,redmon_yolov3_2018,bochkovskiy_yolov4_2020,noauthor_ultralyticsyolov5_2022, li_yolov6_2022,wang_yolov7_2023, jocher_yolo_2023, wang_yolov9_2024, wang_yolov10_2024, jocher_yolo_2023,khanam_yolov11_2024,tian_yolov12_2025} series, SSD \cite{leibe_ssd_2016,fu_dssd_nodate}, RetinaNet \cite{lin_focal_2018}) and two-stage models (e.g., Faster R-CNN \cite{ren_faster_2017}, Cascade R-CNN \cite{cai_cascade_2018}). Single-stage models are known for their speed and suitability for real-time applications, while two-stage models are recognized for their high accuracy, making them ideal for scenarios requiring precise object localization. In multispectral object detection, these models can be enhanced to integrate visible and infrared multispectral information, thereby improving detection performance and demonstrating greater robustness in complex environments such as low-light and low-visibility conditions.

The development of multispectral object detection models typically involves several steps: data preparation, model selection, training, evaluation, and fine-tuning. Once trained, these models are deployed in real-world systems to achieve automated multispectral object detection. As technology advances, more research is focusing on improving detection performance through methods like transfer learning and model fusion. For instance, incorporating attention mechanisms and multispectral feature fusion modules can significantly enhance a model's adaptability and detection accuracy when dealing with multispectral data. These advancements indicate that deep learning-based object detection models have broad application prospects in multispectral detection, offering new possibilities for task automation in complex environments.

\subsection{Multispectral datasets}
Multispectral datasets are essential for research in multispectral object detection, image fusion, and semantic segmentation. With the continuous development of multispectral imaging technologies, several classic datasets have become key tools for evaluating the performance of multispectral algorithms. For example, the KAIST \cite{choi_kaist_2018} and FLIR \cite{zhang_multispectral_2020} datasets, commonly used as benchmarks in multispectral object detection, provide rich pairs of visible and infrared images across various illumination conditions and complex scenarios. The LLVIP \cite{jia_llvip_2021} dataset focuses on visible-infrared paired images under low-light conditions, making it a valuable resource for low-light vision research. Additionally, the M3FD \cite{liu_multispectral_2016} and VEDAI \cite{razakarivony_vehicle_2016} datasets are widely used in multispectral object detection studies. Their diverse image data and detailed annotation information have driven continuous progress in related technologies. Some of the datasets used in this paper's experiments also come from the aforementioned open-source works. In the fields of semantic segmentation and image fusion, the FMB dataset\cite{liu_multi-interactive_2023}, SUNRGBD dataset \cite{song_sun_2015}, and DynamicEarthNet \cite{toker_dynamicearthnet_nodate} dataset offer multimodal data for outdoor, indoor, and satellite scenes, supporting pixel-level semantic segmentation and image fusion tasks. The diversity and complexity of these datasets provide rich resources for research in multispectral object detection, image fusion, and semantic segmentation, promoting the widespread application of multispectral technologies across different fields.

In recent years, the scale and diversity of multispectral datasets have continuously expanded, significantly advancing multispectral object detection technologies. For instance, the DAMSDet \cite{guo_damsdet_nodate} method introduces a dynamic adaptive multispectral detection transformer, which enhances multispectral object detection performance through a modality competition query selection strategy and a multispectral deformable cross-attention module. These research developments show that multispectral datasets not only provide rich multimodal data resources for multispectral object detection but also facilitate the application and development of related technologies in complex environments. This paper focuses on multispectral object detection tasks, aiming to improve detection robustness and accuracy by integrating visible and infrared image information from multispectral datasets.

\subsection{Multispectral feature fusion}
Multispectral feature fusion is a critical component of multispectral object detection, enhancing image information by integrating data from different spectral sensors. Deep learning-based fusion methods, especially those incorporating attention mechanisms and iterative learning strategies, have significantly improved fusion efficiency and robustness. As shown in the lower part of Figure 1, these methods include early fusion \cite{sun_investigating_2019,li_image_2013,li_detail_2021}, mid-level fusion \cite{qingyun_cross-modality_2022,xue_maf-yolo_2021}, mid-to-late fusion \cite{zhang_learning_2018}, late fusion \cite{noauthor_adopting_nodate}, and score fusion \cite{noauthor_adopting_nodate}, each with its unique advantages and applicable scenarios. Early fusion integrates data at the raw data level, capturing complementary information between different modalities from the start. Mid-level fusion, conducted after feature extraction, enhances feature representation. Mid-posterior fusion combines the characteristics of mid-level and late fusion by first fusing features and then performing object detection, thereby improving detection accuracy and robustness. Late fusion and score fusion are two additional effective fusion strategies. Late fusion integrates detection features after each modality has independently completed feature extraction for object detection. This allows for independent evaluation of detection performance across modalities and combines results through specific strategies to boost overall detection performance. Score fusion focuses on detection scores from each modality during the detection process, integrating these scores through weighted averaging, maximum selection, etc., to produce final results. With the development of deep learning technologies, these fusion methods have shown great potential in multispectral image fusion, particularly in handling complex scenes and improving detection accuracy. The framework proposed in this paper encompasses these five fusion modes and combines them with iteratively cross-attention-guided feature fusion to enhance model performance and improve multispectral feature fusion and detection efficacy. Specific details are described in Section \ref{sec:method}.

\begin{figure*}[htbp]
    \centering
    \includegraphics[width=0.9\textwidth]{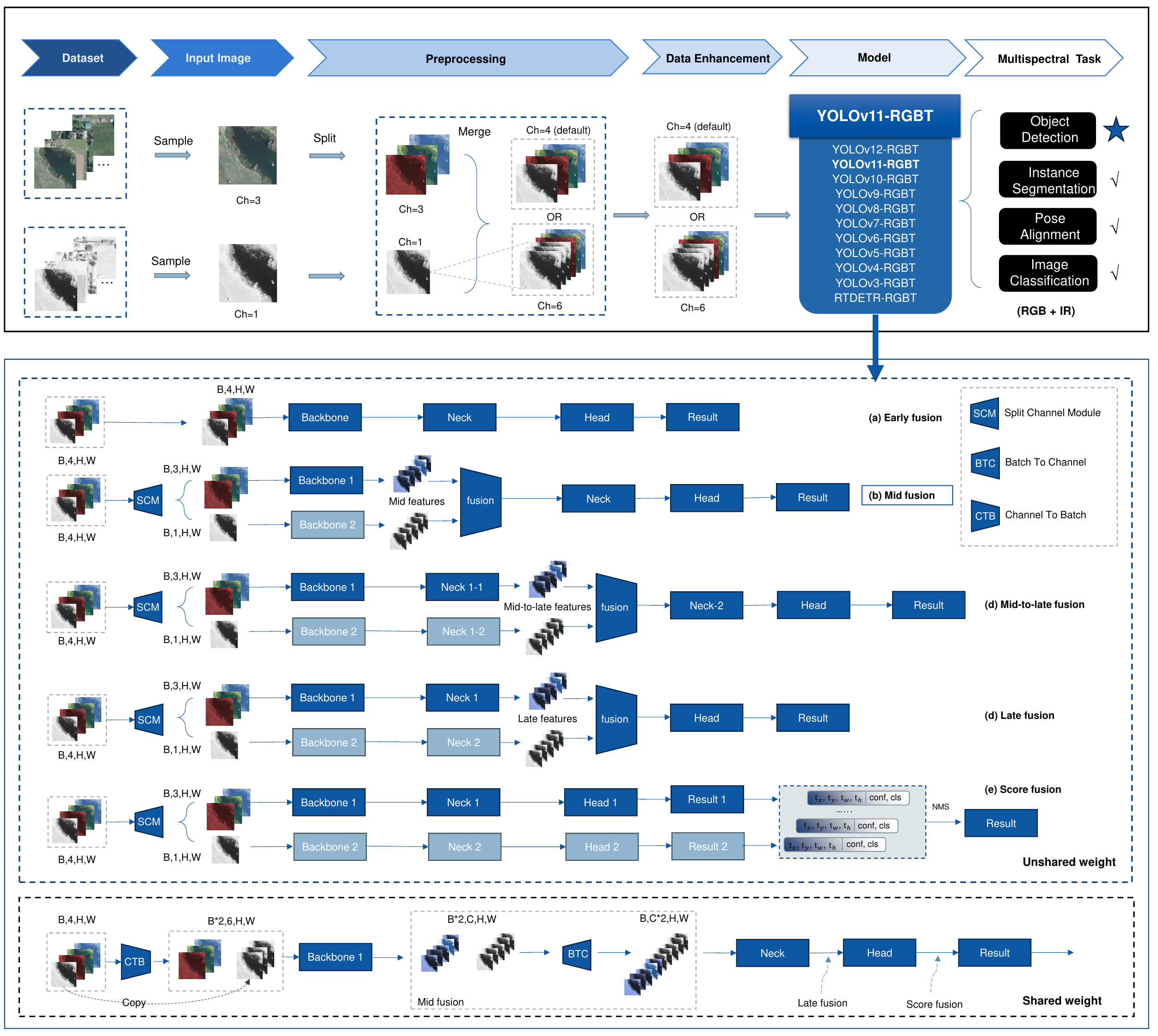}
    \caption{The overall architecture of the YOLOv11-RGBT.}
    \label{fig:fig1}
\end{figure*}

\section{Methodology}
\label{sec:method}
% \lipsum[8] \cite{kour2014real,kour2014fast} and see \cite{hadash2018estimate}.
\subsection{The overall framework of the YOLOv11-RGBT}
This paper presents YOLOv11-RGBT, an integrated framework for multispectral image tasks, based on YOLOv11 \cite{khanam_yolov11_2024}. As shown in Figure \ref{fig:fig1}, it handles multispectral images with RGB and thermal (infrared) data, focusing on improving various multispectral computer vision tasks, particularly multispectral object detection.

\textbf{Model Architecture and Task Execution:} YOLOv11-RGBT's key strength lies in its flexible and efficient architecture supporting YOLOv11's RGBT tasks and other models like YOLOv3-YOLOv12 \cite{redmon_you_2016,redmon_yolo9000_2016,redmon_yolov3_2018,bochkovskiy_yolov4_2020,noauthor_ultralyticsyolov5_2022, li_yolov6_2022,wang_yolov7_2023, wang_yolov9_2024, wang_yolov10_2024, jocher_yolo_2023,khanam_yolov11_2024,tian_yolov12_2025} , RT-DETR \cite{zhao_detrs_2024}, and PP-YOLOE \cite{xu_pp-yoloe_2022} for multispectral detection. The framework comprises three main components: a backbone for feature extraction, a neck for feature processing and fusion, and a head for task execution. This modular design ensures adaptability to diverse applications while maintaining high performance.

\textbf{Data Processing and Augmentation:} Data preprocessing and augmentation are crucial for YOLOv11-RGBT's performance. During preprocessing, multispectral images are standardized and normalized to meet the model's input requirements. Data augmentation techniques like rotation, scaling, and cropping enhance data diversity, improving the model's generalization and adaptability. This process lays a solid foundation for extracting high-quality features from multispectral data.

\textbf{Multispectral Feature Fusion Patterns:} YOLOv11-RGBT supports five fusion modes, including early, mid-level, mid-posterior, late, and score fusion, as well as weight-sharing modes. These innovative combinations of RGB and thermal data boost the model's performance in multispectral environments. By enhancing understanding of multispectral data and improving detection accuracy in complex scenarios, YOLOv11-RGBT effectively utilises multispectral data, providing a powerful tool for multispectral image tasks, especially object detection, and delivering outstanding performance in these tasks.

\subsection{Comparison of multispectral feature mid-fusion strategies}
While some studies indicate that early fusion is more effective for multispectral image fusion tasks \cite{garzelli_multispectral_2018, zhang_rethinking_2024}, mid-level fusion strategies are widely adopted in multispectral object detection \cite{qingyun_cross-modality_2022,noauthor_adopting_nodate,tang_piafusion_2022, zhou_improving_2020,yan_cross-modality_2023}. Our experiments also confirm that mid-level fusion is superior in most scenarios. Consequently, this paper primarily focuses on mid-level fusion strategies.

\begin{figure*}[htbp]
    \centering
    \includegraphics[width=0.9\textwidth]{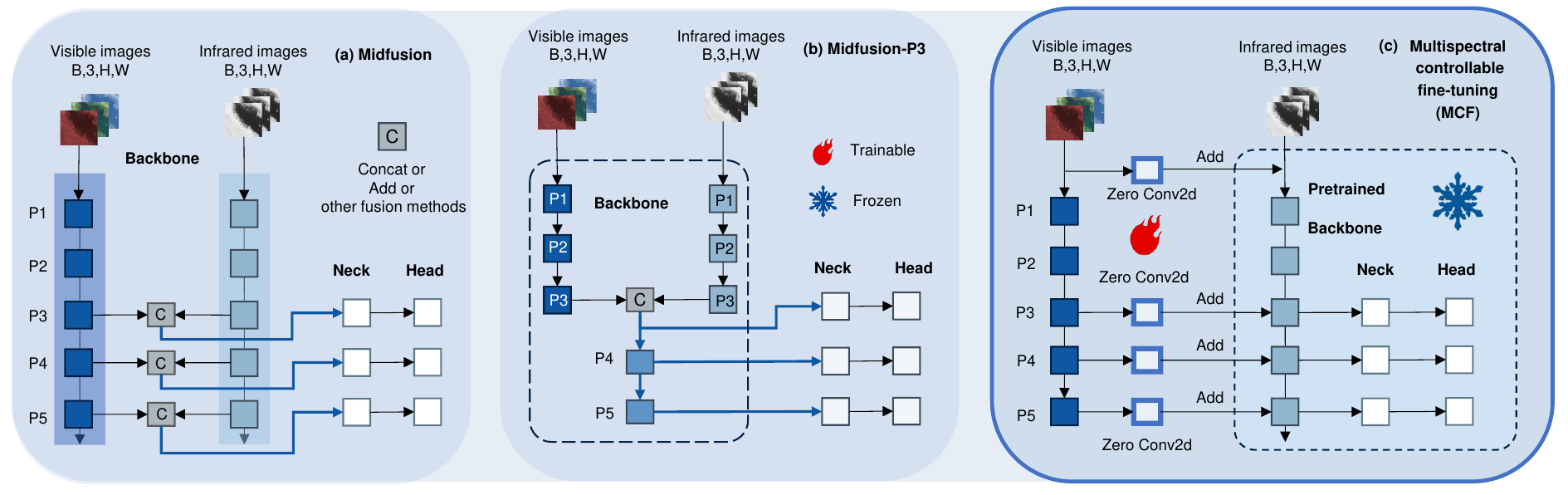}
    \caption{The comparison of multi-spectral intermediate fusion methods for single-stage models.}
    \label{fig:fig2}
\end{figure*}

Three distinct mid-level fusion strategies corresponding to different single-stage multispectral object detection methods are illustrated in our figures. First, Figure \ref{fig:fig2}(a) depicts the conventional mid-level fusion approach. Here, visible and infrared images undergo feature extraction via separate backbones. The resulting feature maps are fused in the neck component using methods like Concat or Add, before being passed to the head for detection output. Fusion typically occurs from the P3 to P5 stages \cite{qingyun_cross-modality_2022,noauthor_adopting_nodate,tang_piafusion_2022}, with some cases involving fusion across all backbone stages \cite{ zhou_improving_2020,yan_cross-modality_2023} (including the dashed parts). Despite leveraging features from multiple levels, this method may introduce interfering information and lead to performance degradation. Moreover, multispectral feature fusion differs from multimodal feature fusion. Many multispectral object detection datasets have aligned features, and multi-level fusion can cause redundancy.

Figure \ref{fig:fig2} (b) presents our proposed P3 mid-level fusion strategy. Fusion occurs at the specific P3 layer, as earlier fusion may not allow sufficient feature extraction. After feature maps from visible and infrared images are extracted by the backbone, they are passed to the neck. At the P3 layer, the feature maps from both modalities are concatenated and processed by a trainable module. This approach effectively utilizes P3 layer features, improving detection accuracy and performance while reducing model parameters and computations.

The P3 fusion lightweight the model by reducing feature fusion nodes, but it is not universally effective across all scenarios. To address this, we propose the multispectral controllable fine-tuning (MCF) strategy shown in Figure \ref{fig:fig2} (c) inspired by ControlNet \cite{zhang_adding_2023}. First, a detection model with excellent performance is trained using infrared images and then frozen to retain pretrained feature representations. Feature maps from visible images are fused with those from infrared images via a Zero Conv2d layer, which is a trainable 2D convolution with initial zero weights. This design allows for controlled fine-tuning of features from different modalities, enhancing model performance stably while utilizing pretrained model knowledge. If a pure visible light model outperforms infrared images (as in the VEDAI dataset), the visible light model can be frozen for fine-tuning. In our experiments, except for the VEDAI and M3FD datasets, we conducted multispectral controllable fine-tuning using models pretrained on infrared images across four datasets. Additionally, while this method primarily introduces information from spectral images, it can also incorporate text, point cloud, or depth data for multimodal object detection. However, this paper focuses on multispectral object detection, and readers are encouraged to explore other methods independently.

\subsection{Multispectral controllable fine-tuning (MCF) strategy}
\begin{figure*}[htbp]
    \centering
    \includegraphics[width=0.9\textwidth]{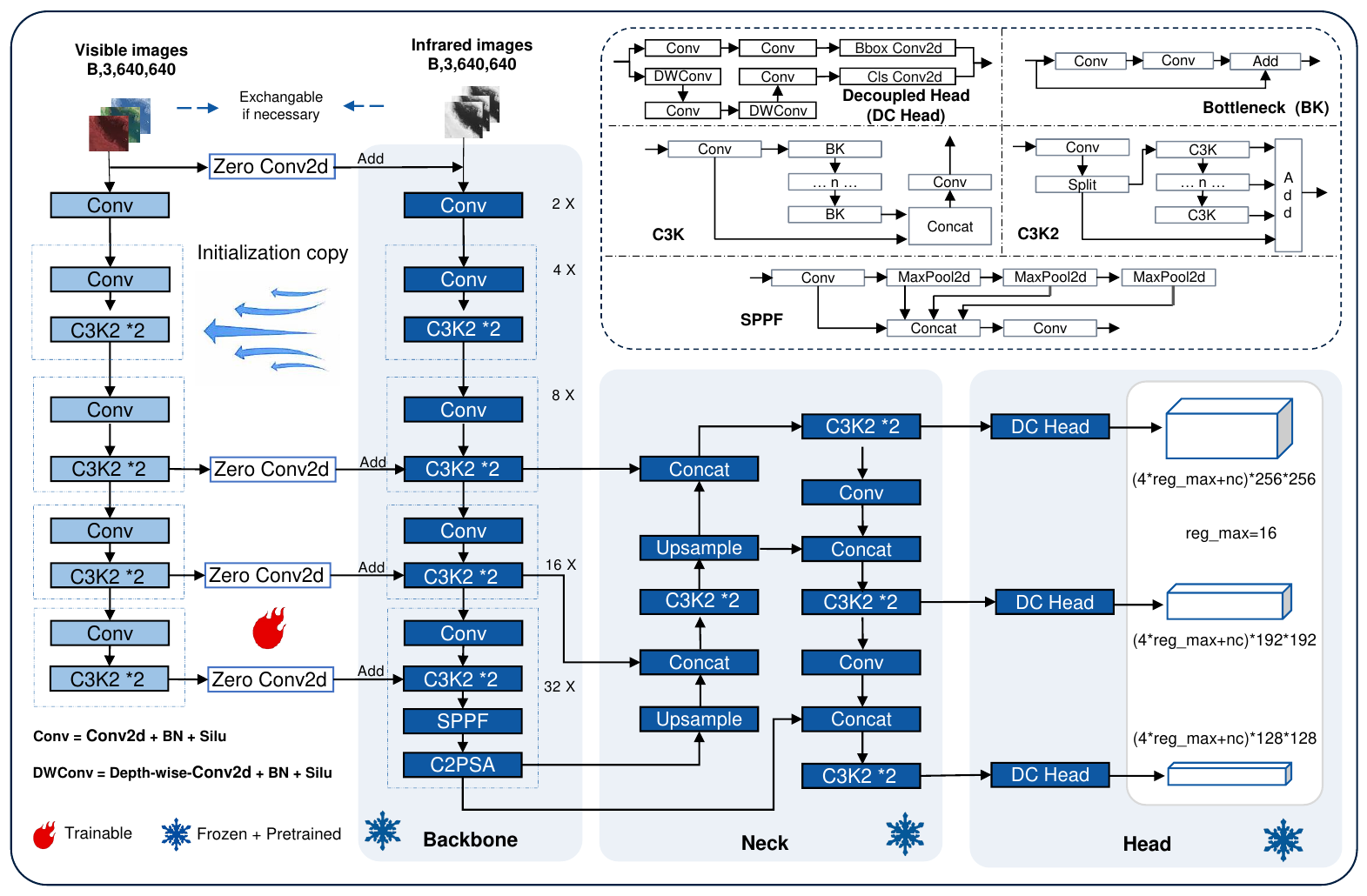}
    \caption{The overall architecture of the YOLOv11-RGBT-MCF.}
    \label{fig:fig3}
\end{figure*}

Figure \ref{fig:fig3} illustrates the overall network architecture of multispectral controllable fine-tuning (MCF) strategy, we embedded it into YOLOv11 as an example and named YOLOv11-RGBT-MCF, which comprises two parts: the frozen component and the Multispectral Controllable Fine-tuning (MCF) component. The frozen component is based on the YOLOv11 base model pretrained on COCO \cite{lin_microsoft_2015} dataset and is divided into three parts: Backbone, Neck, and Head. The Backbone is responsible for extracting image features and consists of multiple convolutional layers (Conv) and C3K2 modules. These modules extract image features from shallow to deep levels. The Neck component, which includes feature fusion, upsampling, and SPPF modules, integrates feature information across different scales to generate more comprehensive feature representations. The Head component, composed of multiple DC Head modules, each corresponding to detection outputs at different scales, enables multiscale object detection. Specific details of these modules are shown in the upper right corner. The Conv module consists of a 2D convolutional layer, a BN (BatchN) layer, and a Silu activation function. The C3K2 module consists of a 2D convolutional layer and a bottleneck layer. These designs enable the network to learn more features through multi-branch learning during training, thereby enhancing detection performance.

The MCF strategy enhances the base model by fine-tuning it with visible light image features. This is achieved using a Zero Conv2d layer, which is a trainable 2D convolutional layer with initial zero weights. The Zero Conv2d layer allows for controlled fusion of visible light features with infrared features from the frozen model, enabling targeted fine-tuning of the single-modal model. Unlike ControlNet , which often fuses features in later stages like the Neck and Head, our MCF strategy focuses on mid-level fusion. This approach is more suitable for multispectral object detection models and allows for more effective information integration.

\subsection{Multispectral transfer training principle in YOLOv11-RGBT}
When conducting transfer training for YOLOv11-RGBT, the core principle is to load the pre-trained model weights from the COCO \cite{lin_microsoft_2015} dataset into the multispectral model architecture. If the multispectral model structure is identical to the pre-trained model, the corresponding weights can be directly copied, ensuring a seamless parameter transfer. However, when encountering structural discrepancies, we utilize several effective strategies to ensure model compatibility and performance. Specific details can be found in the repository code.

For instance, in cases of inconsistent channels, channel averaging or copying can be applied to achieve uniformity, laying the foundation for subsequent training. Additionally, inserting 1×1 convolutional layers can adjust channel consistency, enabling the model to better process multispectral data and integrate information from different spectra, thereby enhancing target detection capabilities. Taking Midfusion as an example, its transfer training process involves replicating the YOLOv11 backbone into separate backbones for visible and infrared images. The neck and head components can then be directly copied, rapidly completing the transfer training and improving detection performance and generalisation in various scenarios.

\subsection{Loss function}
The loss function of YOLOv11-RGBT is consistent with YOLOv11 and is divided into 3 parts: Distribution focal loss $L_{\text{all}}$ , object classification loss $L_{\text{cls}}$, and object localisation loss $ L_{\text{loc}}$. The loss function formula was as follows:
\begin{equation} % 公式 1
L_{\text{all}} = \lambda_{\text{dfl}} L_{\text{dfl}} + \lambda_{\text{cls}} L_{\text{cls}} + \lambda_{\text{loc}} L_{\text{loc}}
\end{equation}

Where $L_{\text{all}}$  contains three parts and $\lambda$ is a hyperparameter representing the weights of each part. These weights can be adjusted before training according to actual conditions. In this paper, the weights for the three parts are 1.0, 0.5, and 0.05, respectively.

The classification loss  $L_{{cls}}$ utilises binary cross-entropy (BCE) loss, expressed as:
\begin{equation} % 公式 2
\begin{split}
L_{\text{cls}} =& -\sum_{I=0}^{K \times K} \Big[ I_{ij}^{\text{obj}} \sum_{c \in \text{classes}} \Big\{ P_i^j(c) \log \Big[ P'_i{}^j(c) \Big] \\
&+ \Big[ 1 - P_i^j(c) \Big] \log \Big[ 1 - P'_i{}^j(c) \Big] \Big\} \Big]
\end{split}
\end{equation}
Here, K*K can take three values depending on the image size (e.g., for the image size of 640*640, they were 20*20, 40*40, 80*80), representing the grid numbers on three different scale feature maps. output by YOLOv11-RGBT $I_{ij}^{\text{obj}}$ indicates whether the $j^{th}$ prior box in the $i^{th}$ grid has a predicted target (1 for yes, 0 for no). The c represents the target category, and $ P_i^j(c)$ and $P'_i{}^j(c)$ are the probabilities of the target belonging to a certain category in the ground truth and prediction, respectively.

The object localization loss employs CIOU Loss and incorporates three geometric parameters: overlap area, center point distance, and aspect ratio. These parameters are instrumental in refining the predicted box to better align with the ground truth box, thereby enhancing regression accuracy. The formula for the loss function is as follows:

\begin{equation} % 公式 3
L_{\text{loc}} = 
\begin{cases}
1 - \text{IoU} + \frac{\rho^2(b_{\text{pred}}, b_{\text{gt}})}{c^2} + \alpha v \\
\alpha = v / (1 - \text{IoU} + v) \\
v = \frac{4}{\pi^2} \left( \arctan \frac{w_{\text{gt}}}{h_{\text{gt}}} - \arctan \frac{w}{h} \right)^2
\end{cases} 
\end{equation}

Where, $\rho^2(b_{pred},b_{gt})$ represents the Euclidean distance between the center points of the predicted box and the ground truth box, c is the diagonal distance of the smallest closed bounding box that could contain both the predicted box and the ground truth box, and $w_{gt}$, $h_{gt}$ are the width and height of the ground truth box, while w, h are the width and height of the predicted box.

$L_{dfl}$ is the Distribution Focal Loss (DFL) aimed at quickly focusing the network on values near the annotated positions and maximizing their probabilities. The expression is:
\begin{equation}
L_{\text{dfl}} = \sum_{I=0}^{K \times K} \sum_{p=0}^3 I_{ij}^{\text{obj}} \cdot \text{DFL}(s_i, s_{i+1}) 
\end{equation}
 
Here,  K*K  is consistent with formula 4, and p represents the four predicted coordinate values. DFL regresses the predicted boxes in a probabilistic way, requiring setting a hyperparameter reg\_max in advance, default reg\_max is 16. At this point, the output channel of this branch of the network is 64 = 4 * reg\_max. Before that, 16 fixed reference values A: [0, 1, 2, ..., 15], are set, corresponding to each position of  reg\_max . For these reg\_max numbers, the softmax function is utilized for discretization, treating it as a 16-class classification. Cross-entropy loss is employed for calculating the loss, as shown in the formula:
\begin{equation} % 公式 5
\text{DFL}(S_i, S_{i+1}) = - \Big[ (y_{i+1} - y) \log(S_i) + (y - y_i) \log(S_{i+1}) \Big] 
\end{equation}

The target position coordinates obtained in the feature map generally do not fall on specific grid corners, but labels need to be integers. Taking the prediction $x_{min}$ as an example, its true value is y, where the left integer is $y_{i}$ and the right integer is $y_{i+1}$.  The $(y_{i+1}-y)$  and $(y-y_i)$ correspond to the weights of the distances from the true value, $S_i$ and $S_{i+1}$ correspond to the predicted values of  $y_i$ and $y_{i+1}$, respectively.

\section{Experiments}
\label{experiments}
The experimental platform, datasets, and details for this study are presented in \cref{experiments:experimental platform,experiments:experimental datasets,experiments:experimental details}, with additional details available in the code. Sections \ref{experiments:experiments FLIR} and \ref{experiments:experiments LLVIP} aim to show that mid-term multispectral fusion can sometimes reduce model detection performance in certain scenarios, while also demonstrating the effectiveness and feasibility of the proposed MCF method. Section \ref{experiments:experiments M3FD} focuses on proving the framework's effectiveness and feasibility in typical multispectral detection tasks, as well as the practicality of multispectral transfer learning.

% \lipsum[8] \cite{kour2014real,kour2014fast} and see \cite{hadash2018estimate}.
\subsection{Experimental platform and related indicators}
\label{experiments:experimental platform}
\begin{table}[htbp]
    \centering
    \caption{Experimental platform}
    \label{tab:experimental_platform}
    \begin{tabular}{lcc}
        \toprule
        \textbf{Platform} & \textbf{LLVIP} & \textbf{Other datasets} \\
        \midrule
        CPU & Intel(R) Xeon(R) Gold 5418Y & Intel(R) Xeon(R) CPU E5-2680 \\
        GPU & NVIDIA GeForce RTX 4090 & NVIDIA GeForce RTX 3090 \\
        The operating system & Ubuntu20.04 & Ubuntu20.04 \\
        Deep learning framework & Pytorch 2.2 & Pytorch 1.12 \\
        \bottomrule
    \end{tabular}
\end{table}
Table \ref{tab:experimental_platform} illustrates the experimental platform. Evaluation of network performance was primarily dependent on the mAP (mean average precision) during training and the performance of the trained network in the verification set. To quantify the detection results, precision (P), recall (R), and mAP[57] were used as performance evaluation indices. This is the expression for P and R:
\begin{equation} % 公式 6
    R = \frac{TP}{TP + FN} 
\end{equation}

\begin{equation} % 公式 7
    P = \frac{TP}{TP + FP} 
\end{equation}

True positives (TP): the number of positive samples that the classifier correctly identified as positive samples. True negatives (TN): the number of samples that are truly negative and are divided by the classifier into negative samples. False positives (FP): the number of samples that are truly negative but are misclassified by the classifier as positive. False negatives (FN): the number of positive samples that are incorrectly classified as negative by a classifier.

Average precision (AP) is the region bounded by the P-R curves. In general, the higher the AP value, the better the classifier. The mAP is a comprehensive measure of the average accuracy of detected targets. The mAP is used to calculate the average value of each category's APs individually. These expressions describe AP and mAP:
\begin{equation} % 公式 8
    AP_i = \int_{0}^{1} P_i(R_i) dR_i = \sum_{k=0}^{n} P_i(k) \Delta R_i(k) 
\end{equation}

\begin{equation} % 公式 9
    mAP = \frac{1}{C} \sum_{c=1}^{C} AP_i 
\end{equation}

\subsection{Experimental datasets}
\label{experiments:experimental datasets}
\begin{figure}[htbp]
	\centering
	\begin{subfigure}{0.325\linewidth}
		\centering
		\includegraphics[width=0.9\linewidth]{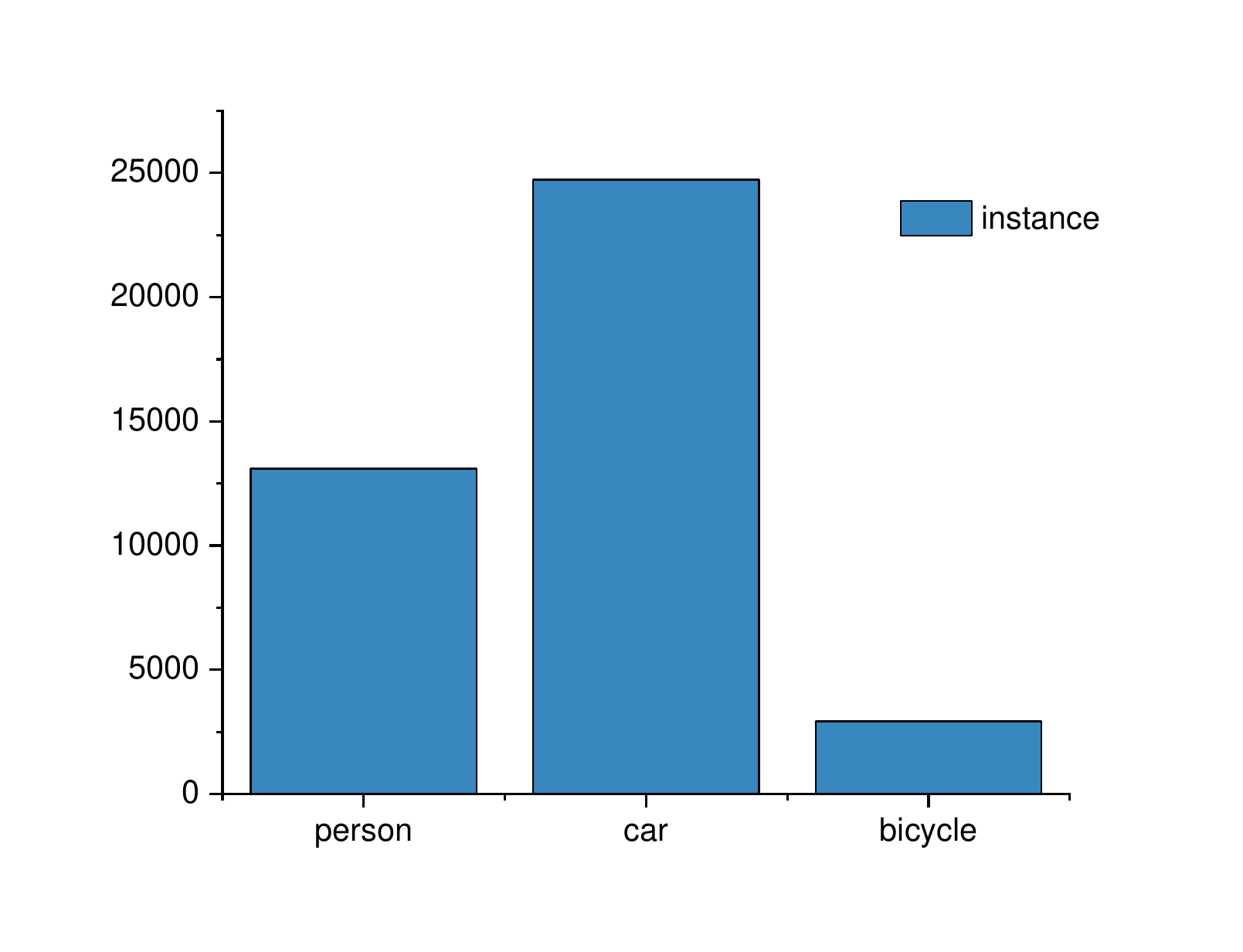}
		\caption{FLIR}
		\label{FLIR}%文中引用该图片代号
	\end{subfigure}
	\centering
	\begin{subfigure}{0.325\linewidth}
		\centering
		\includegraphics[width=0.9\linewidth]{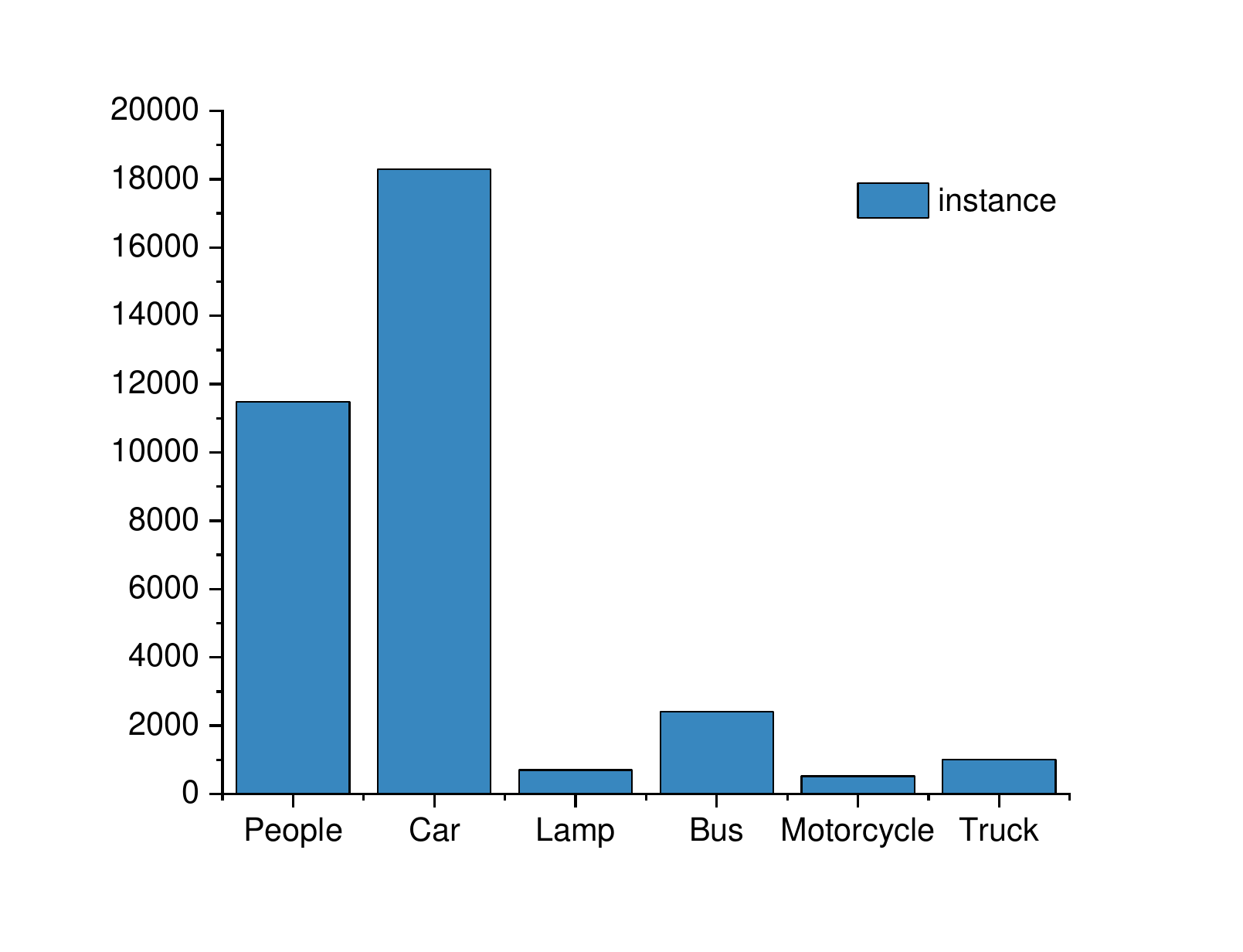}
		\caption{M3FD}
		% \label{chutian3}%文中引用该图片代号
	\end{subfigure}
	\centering
	\begin{subfigure}{0.325\linewidth}
		\centering
		\includegraphics[width=0.9\linewidth]{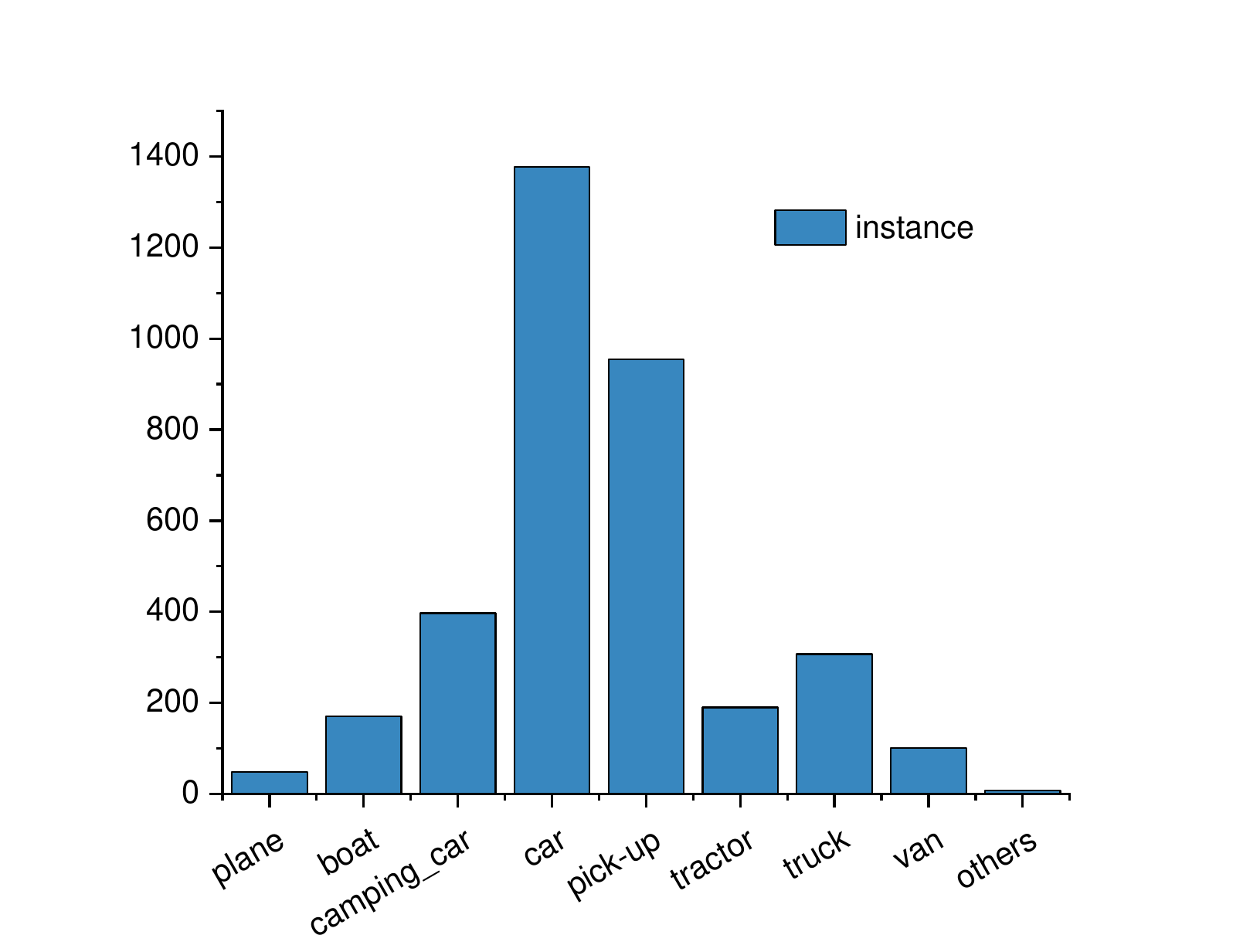}
		\caption{VEDAI}
		\label{VEDAI}%文中引用该图片代号
	\end{subfigure}
	\caption{Distribution of the number of objects in each dataset. The horizontal axis is the category name, and the vertical axis is the count of each categorie. (a) FLIR; (b) M3FD; (c) VEDAI.}
	\label{fig4}
\end{figure}

We utilized five open-source multispectral object detection datasets to verify the effectiveness, feasibility, and generalization ability of our detection system and algorithm in complex scenarios. All images from these datasets were resized to 640×640 before being input into the network. These datasets can be downloaded from their official websites or via the links in the GitHub introduction document. Below is a brief introduction to each dataset:

\textbf{FLIR \cite{zhang_multispectral_2020}:} Captured using infrared thermographic cameras, this dataset primarily annotates three categories: pedestrians, cars, and bicycles. With an image size of 640×512 pixels, it is pre-registered and consists of 4,124 training pairs and 1,013 testing pairs. It is commonly used for object detection, especially in complex scenarios like night-time and low-light conditions. The category distribution is shown in Figure \ref{fig4}(a).

\textbf{M3FD \cite{liu_target-aware_2022}:} Collected with dual-optical cameras and infrared sensors, it contains 4,200 image pairs and annotates six categories, including humans, cars, and trucks. Widely used in image fusion and object detection tasks, 3,360 images were selected as the training set and 840 as the validation set. Its category distribution is illustrated in Figure \ref{fig4}(b).

\textbf{KAIST \cite{choi_kaist_2018}:} The original KAIST dataset was captured using visible and long-wave infrared (LWIR) sensors. This study employs the version readjusted by Li, which only annotates pedestrian targets. Mainly used for pedestrian detection tasks, it comprises 8,956 training pairs and 2,252 validation pairs, making it suitable for multispectral pedestrian detection research.

\textbf{LLVIP \cite{jia_llvip_2021}:} Acquired with visible and infrared cameras, it consists of 15,488 image pairs and annotates the pedestrian category. With an image size of 640×512 pixels, it is pre-registered and divided into 12,025 training pairs and 3,463 testing pairs. It is primarily used for low-light-vision tasks such as image fusion and object detection. 

\textbf{VEDAI \cite{razakarivony_vehicle_2016}:} Captured via aerial visible and infrared cameras, it includes approximately 1,050 image pairs with a size of around 640×640 pixels. Pre-registered and without official fixed splits, it was divided into training and testing sets at a ratio of 8:2. Mainly used for object detection tasks, its category distribution is shown in Figure \ref{fig4}(c).

\subsection{Implementation details}
\label{experiments:experimental details}
Experiments in this paper were conducted on two open-source frameworks: our YOLOv11-RGBT and MMDetection. We selected multiple models, including YOLOv3-YOLOv12 and RT-DETR, for comparative experiments. To boost result reproducibility, hyperparameters were barely altered and kept consistent across model training. When experimenting with the aforementioned datasets, the general settings were as follows: a batch size of 16 and a model input resolution of 640×640. If GPU memory was insufficient, the batch size was reduced to 8. For MMDetection, training involved 3 repeated batches and 30 epochs, with ResNet50 as the backbone. Models in other frameworks were trained for 300 epochs. To speed up training, workers were set to 8 where possible. Below are brief model introductions:

\textbf{YOLOv3 \cite{redmon_yolov3_2018}:} YOLOv3 is a one-stage object detection model in the YOLO series. By incorporating multi-scale feature maps and utilizing a larger network structure, YOLOv3 improves the accuracy and detection capability for small objects.

\textbf{YOLOv4 \cite{bochkovskiy_yolov4_2020}:} Upgraded from YOLOv3 with CSPDarknet53 backbone, Mish activation, and SPP module for enhanced speed and precision.

\textbf{YOLOv5 \cite{noauthor_ultralyticsyolov5_2022}:} YOLOv5 is a significant version in the YOLO series, featuring a lightweight network structure and model compression techniques. While maintaining high accuracy, it notably enhances detection speed and model efficiency, making it suitable for mobile devices and embedded systems.

\textbf{YOLOv6 \cite{li_yolov6_2022}:} Developed by Meituan, focuses on industrial applications with efficient decoupled heads and reparameterization techniques.

\textbf{YOLOv7 \cite{wang_yolov7_2023}:} YOLOv7 also employs extensive reparametrization techniques and introduces trainable bag-of-freebies methods to significantly improve detection accuracy in real-time without increasing inference costs. Upon release, it surpassed all known object detectors in terms of both speed and accuracy.

\textbf{YOLOv8 \cite{jocher_yolo_2023}:} YOLOv8 is a derived model based on enhancements and optimizations from YOLOv5, aiming to further enhance object detection performance and effectiveness. These improvements involve adjustments in network structure, training strategies, data augmentation, with the most significant change being the transition to an Anchor-Free paradigm.

\textbf{YOLOv9 \cite{wang_yolov9_2024}:} Incorporates GELAN modules and deep supervision for better gradient flow and convergence in resource-constrained systems.

\textbf{YOLOv10 \cite{wang_yolov10_2024}:} Introduces uniform double assignment strategy for NMS-free training and a lightweight classification head for efficiency.

\textbf{YOLOv11 \cite{khanam_yolov11_2024}:} Focuses on computational efficiency with C3k2 and C2PSA modules for improved feature extraction without accuracy loss.

\textbf{YOLOv12 \cite{tian_yolov12_2025}:} Optimized from YOLOv8 with attention mechanisms for better feature extraction but slightly reduced generalization.

\textbf{RT-DETR \cite{zhao_detrs_2024}:} Based on Transformer architecture, removes traditional NMS steps for reduced computational complexity and faster inference.

\textbf{RetinaNet\cite{lin_focal_2018}:} RetinaNet is a single-stage object detection model that addresses class imbalance issues in object detection using a feature pyramid network and Focal Loss. It achieves efficient and accurate object detection, particularly excelling in handling small objects.

\textbf{Faster R-CNN \cite{ren_faster_2017}:} Faster R-CNN is a two-stage object detection model that introduces a Region Proposal Network (RPN) to generate candidate regions and utilizes a shared feature extraction network for classification and precise localization. It strikes a good balance between accuracy and speed.

\textbf{Cascade R-CNN \cite{cai_cascade_2018}:} Cascade R-CNN is an improved two-stage object detection model that cascades multiple R-CNN modules to progressively filter candidate boxes, enhancing object detection accuracy, especially suitable for small object detection and complex scenes.

\subsection{Comparative experiments on FLIR dataset}
\label{experiments:experiments FLIR}
The \cref{tab:flir_rgb_results,tab:flir_ir_results,tab:flir_rgb_ir_results,tab:flir_rgb_ir_results_p3,tab:flir_fine_tuning,tab:flir_model_comparison} present the comparative results of multiple models in the FLIR data set. Table \ref{tab:flir_rgb_results} shows the effects of models trained solely on visible light images, while Table \ref{tab:flir_ir_results} presents results from models trained only on infrared images. Together, they offer a comprehensive evaluation of the latest YOLO models on FLIR. The \cref{tab:flir_rgb_ir_results,tab:flir_rgb_ir_results_p3} show results of models trained with Midfusion and Midfusion-P3 methods. Notably, all models in  \cref{tab:flir_rgb_results,tab:flir_ir_results,tab:flir_rgb_ir_results,tab:flir_rgb_ir_results_p3} were trained without pre-trained weights. A row-by-row analysis reveals that most multispectral-trained models in \cref{tab:flir_rgb_ir_results,tab:flir_rgb_ir_results_p3} outperform the visible-light-only models in Table \ref{tab:flir_rgb_results}, but few surpass the infrared-only models. This indicates that infrared images dominate in FLIR, as visible light images are less effective than infrared thermal images in harsh conditions like night-time or fog. For example, YOLOv11n-Midfusion improved mAP by 1.10\% over YOLOv11n infrared models, and YOLOv3-Tiny's 3-node fusion model increased mAP 50:95 by 0.91\% compared to infrared-only models. These results confirm the effectiveness of our multispectral models and the superiority of the YOLOv11-RGBT framework.

\begin{figure*}[htbp]
    \centering
    \includegraphics[width=0.6\textwidth]{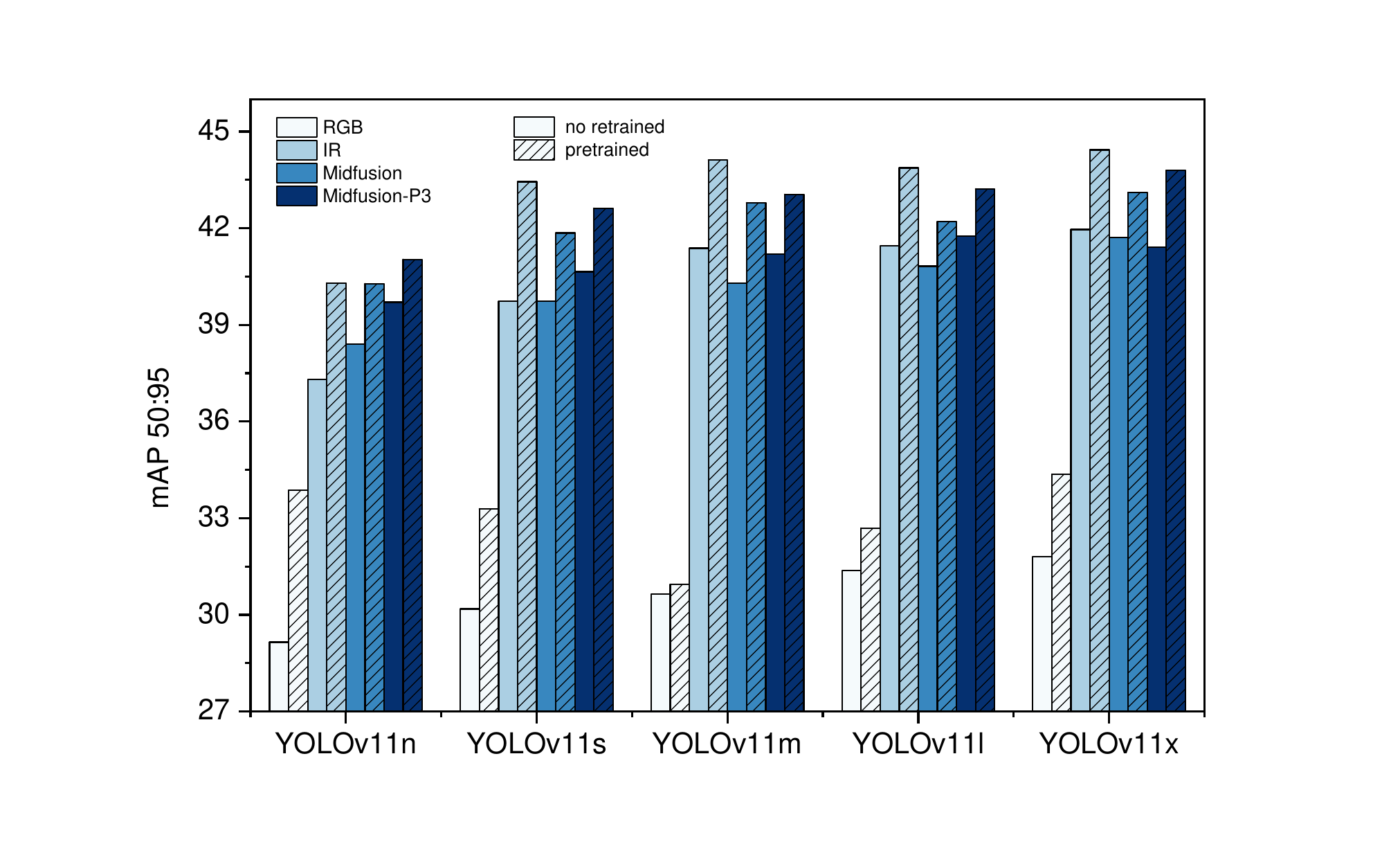}
    \caption{The transfer learning results of several YOLOv11 models after loading COCO-pretrained weights.}
    \label{fig:fig5}
\end{figure*}

\begin{figure*}[htbp]
    \centering
    \includegraphics[width=0.6\textwidth]{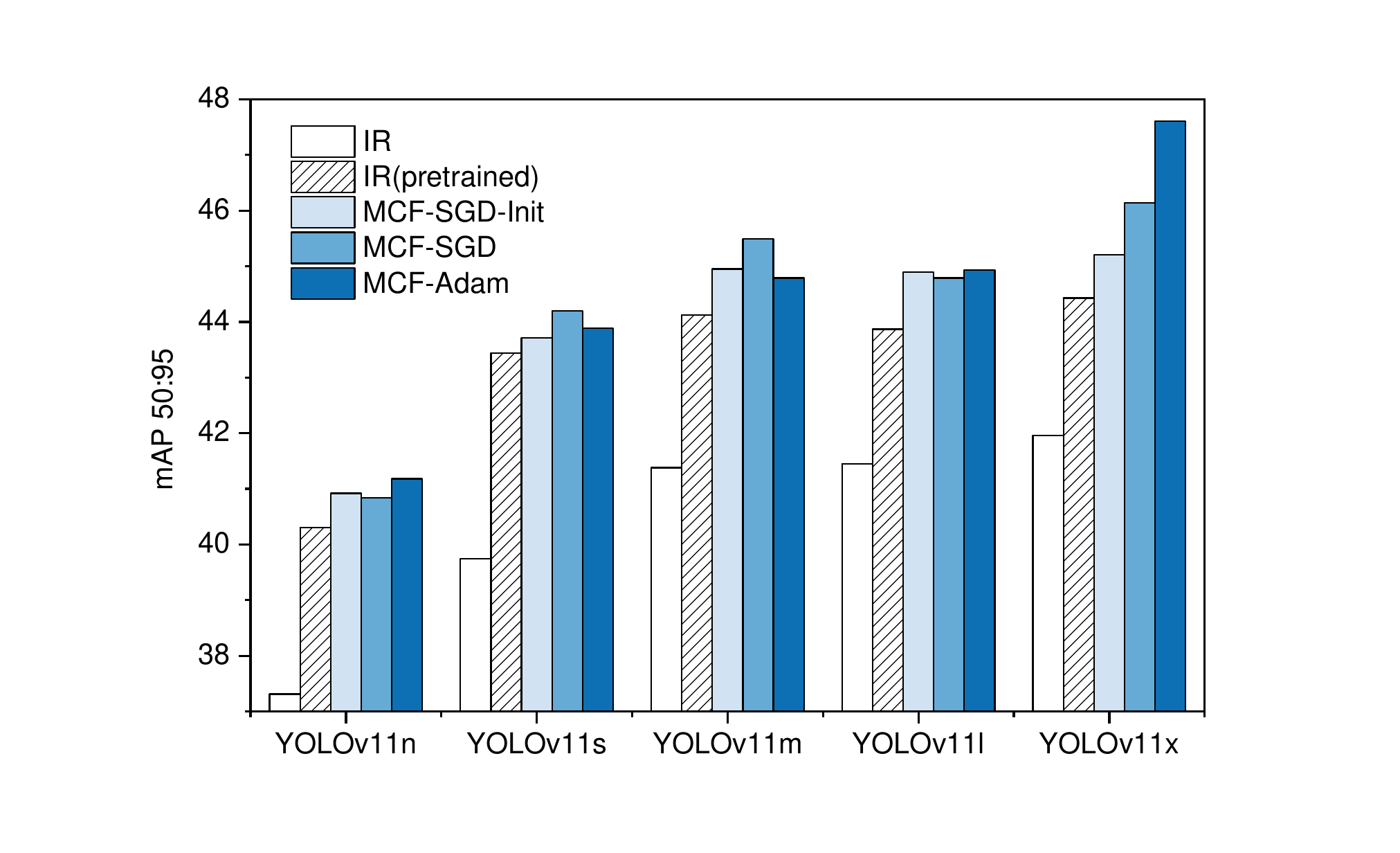}
    \caption{The comparison results of multispectral controllable fine-tuning (MCF) strategy utilized different hyperparameters.}
    \label{fig:fig6}
\end{figure*}

\begin{table}[htbp]
    \centering
    \caption{The comparison results of object detection models on the FLIR dataset using the visible images (RGB)}
    \label{tab:flir_rgb_results}
    {
    \begin{tabular}{@{}lcccccc@{}}
        \toprule
        \textbf{Method} & \textbf{Publication} & \textbf{Year} & \textbf{Params} & \textbf{Flops} & \textbf{AP50} & \textbf{AP} \\
                        &                     &               & \textbf{(M)}    & \textbf{(G)}   & \textbf{(\%)} & \textbf{(\%)} \\
        \midrule
        YOLOv3-Tiny\cite{redmon_yolov3_2018} & arXiv & 2018 & 98.89 & 282.99 & 53.31 & 24.46 \\
        YOLOv3\cite{redmon_yolov3_2018} & arXiv & 2018 & 11.57 & 19.05 & 64.15 & 30.81 \\
        YOLOv4-Tiny\cite{bochkovskiy_yolov4_2020} & arXiv & 2020 & 86.77 & 217.34 & 56.88 & 26.61 \\
        YOLOv4 \cite{bochkovskiy_yolov4_2020} & arXiv & 2020 & 7.39 & 14.56 & 65.16 & 32.26 \\
        YOLOv5n \cite{noauthor_ultralyticsyolov5_2022} & 7.0u & 2020 & 2.39 & 7.18 & 58.83 & 28.25 \\
        YOLOv5s \cite{noauthor_ultralyticsyolov5_2022} & 7.0u & 2020 & 8.70 & 24.05 & 61.52 & 29.62 \\
        YOLOv5m \cite{noauthor_ultralyticsyolov5_2022} & 7.0u & 2020 & 23.91 & 64.36 & 62.77 & 30.43 \\
        YOLOv5l\cite{noauthor_ultralyticsyolov5_2022} & 7.0u & 2020 & 50.70 & 135.29 & 63.91 & 30.15 \\
        YOLOv5x\cite{noauthor_ultralyticsyolov5_2022} & 7.0u & 2020 & 92.70 & 246.91 & 63.14 & 30.44 \\
        YOLOv6n\cite{li_yolov6_2022} & arXiv & 2022 & 4.04 & 11.87 & 61.38 & 29.60 \\
        YOLOv6s\cite{li_yolov6_2022} & arXiv & 2022 & 15.55 & 44.21 & 61.96 & 29.65 \\
        YOLOv6m\cite{li_yolov6_2022} & arXiv & 2022 & 49.59 & 161.55 & 62.36 & 30.56 \\
        YOLOv6l\cite{li_yolov6_2022} & arXiv & 2022 & 105.76 & 391.93 & 62.56 & 30.00 \\
        YOLOv6x\cite{li_yolov6_2022} & arXiv & 2022 & 165.01 & 611.15 & 64.37 & 30.61 \\
        YOLOv7-Tiny\cite{wang_yolov7_2023} & CVPR & 2023 & 42.17 & 132.18 & 61.80 & 30.02 \\
        YOLOv7\cite{wang_yolov7_2023} & CVPR & 2023 & 8.01 & 22.05 & 65.94 & 32.08 \\
        YOLOv8n\cite{jocher_yolo_2023} & 8.3u & 2024 & 2.87 & 8.20 & 59.09 & 28.23 \\
        YOLOv8s\cite{jocher_yolo_2023} & 8.3u & 2024 & 10.62 & 28.65 & 60.88 & 28.91 \\
        YOLOv8m\cite{jocher_yolo_2023} & 8.3u & 2024 & 24.66 & 79.07 & 63.87 & 30.42 \\
        YOLOv8l\cite{jocher_yolo_2023} & 8.3u & 2024 & 41.61 & 165.41 & 62.70 & 30.07 \\
        YOLOv8x\cite{jocher_yolo_2023} & 8.3u & 2024 & 65.00 & 258.13 & 63.27 & 31.01 \\
        YOLOv9t\cite{wang_yolov9_2024} & ECCV & 2024 & 1.91 & 7.85 & 61.10 & 29.30 \\
        YOLOv9s\cite{wang_yolov9_2024} & ECCV & 2024 & 6.95 & 27.39 & 63.91 & 30.80 \\
        YOLOv9m\cite{wang_yolov9_2024} & ECCV & 2024 & 19.23 & 77.56 & 65.94 & 31.43 \\
        YOLOv10n\cite{wang_yolov10_2024} & NeurIPS & 2024 & 2.58 & 8.40 & 58.36 & 28.29 \\
        YOLOv10s\cite{wang_yolov10_2024} & NeurIPS & 2024 & 7.69 & 24.78 & 61.48 & 29.88 \\
        YOLOv10m\cite{wang_yolov10_2024} & NeurIPS & 2024 & 15.72 & 63.98 & 62.21 & 29.90 \\
        YOLOv10l\cite{wang_yolov10_2024} & NeurIPS & 2024 & 24.58 & 127.21 & 60.85 & 29.31 \\
        YOLOv10x\cite{wang_yolov10_2024} & NeurIPS & 2024 & 30.19 & 171.02 & 62.47 & 30.55 \\
        YOLOv11n\cite{khanam_yolov11_2024} & arXiv & 2024 & 2.47 & 6.44 & 60.37 & 29.15 \\
        YOLOv11s\cite{khanam_yolov11_2024} & arXiv & 2024 & 8.99 & 21.55 & 62.15 & 30.18 \\
        YOLOv11m\cite{khanam_yolov11_2024} & arXiv & 2024 & 19.13 & 68.20 & 64.00 & 30.64 \\
        YOLOv11l\cite{khanam_yolov11_2024} & arXiv & 2024 & 24.14 & 87.28 & 63.58 & 31.37 \\
        YOLOv11x\cite{khanam_yolov11_2024} & arXiv & 2024 & 54.24 & 195.46 & 66.23 & 31.80 \\
        YOLOv12n\cite{tian_yolov12_2025} & arXiv & 2025 & 2.45 & 6.48 & 60.99 & 29.54 \\
        YOLOv12s\cite{tian_yolov12_2025} & arXiv & 2025 & 8.83 & 21.53 & 63.25 & 30.50 \\
        YOLOv12m\cite{tian_yolov12_2025} & arXiv & 2025 & 19.21 & 67.75 & 63.86 & 31.50 \\
        YOLOv12l\cite{tian_yolov12_2025} & arXiv & 2025 & 25.17 & 89.42 & 64.70 & 31.81 \\
        YOLOv12x\cite{tian_yolov12_2025} & arXiv & 2025 & 56.38 & 199.83 & 64.65 & 31.70 \\
        RT-DETR-r50\cite{zhao_detrs_2024} & CVPR & 2024 & 40.79 & 130.48 & 56.66 & 26.25 \\
          RetinaNet\cite{lin_focal_2018} & ICCV  &2017 & 36.43 & 61.89 & 45.40 & 18.60 \\
       Faster R-CNN\cite{ren_faster_2017} & NeurIPS  &2015 & 41.36 & 75.51 & 44.80 & 18.90 \\
         Cascade R-CNN\cite{cai_cascade_2018} & TMAMI  &2019 & 69.16 & 103.30 & 41.10 & 17.60 \\
        \bottomrule
    \end{tabular}
    }
\end{table}

\begin{table}[htbp]
    \centering
    \caption{The comparison results of object detection models on the FLIR dataset using the infrared images (IR)}
    \label{tab:flir_ir_results}
    {
    \begin{tabular}{@{}lcccccc@{}}
        \toprule
        \textbf{Method} & \textbf{Publication} & \textbf{Year} & \textbf{Params} & \textbf{Flops} & \textbf{AP50} & \textbf{AP} \\
                        &                     &               & \textbf{(M)}    & \textbf{(G)}   & \textbf{(\%)} & \textbf{(\%)} \\
        \midrule
        YOLOv3-Tiny\cite{redmon_yolov3_2018} & arXiv & 2018 & 98.89 & 282.99 & 67.64 & 34.08 \\
        YOLOv3\cite{redmon_yolov3_2018} & arXiv & 2018 & 11.57 & 19.05 & 77.72 & 41.89 \\
        YOLOv4-Tiny\cite{bochkovskiy_yolov4_2020} & arXiv & 2020 & 86.77 & 217.34 & 71.71 & 37.12 \\
        YOLOv4\cite{bochkovskiy_yolov4_2020} & arXiv & 2020 & 7.39 & 14.56 & 78.16 & 42.44 \\
        YOLOv5n\cite{noauthor_ultralyticsyolov5_2022} & 7.0u & 2020 & 2.39 & 7.18 & 72.18 & 38.07 \\
        YOLOv5s\cite{noauthor_ultralyticsyolov5_2022} & 7.0u & 2020 & 8.70 & 24.05 & 72.92 & 39.45 \\
        YOLOv5m\cite{noauthor_ultralyticsyolov5_2022} & 7.0u & 2020 & 23.91 & 64.36 & 76.19 & 40.85 \\
        YOLOv5l\cite{noauthor_ultralyticsyolov5_2022} & 7.0u & 2020 & 50.70 & 135.29 & 76.41 & 41.67 \\
        YOLOv5x\cite{noauthor_ultralyticsyolov5_2022} & 7.0u & 2020 & 92.70 & 246.91 & 78.03 & 42.07 \\
        YOLOv6n\cite{li_yolov6_2022} & arXiv & 2022 & 4.04 & 11.87 & 74.73 & 39.81 \\
        YOLOv6s\cite{li_yolov6_2022} & arXiv & 2022 & 15.55 & 44.21 & 76.29 & 40.71 \\
        YOLOv6m\cite{li_yolov6_2022} & arXiv & 2022 & 49.59 & 161.55 & 77.36 & 42.62 \\
        YOLOv6l\cite{li_yolov6_2022} & arXiv & 2022 & 105.76 & 391.93 & 78.79 & 43.27 \\
        YOLOv6x\cite{li_yolov6_2022} & arXiv & 2022 & 165.01 & 611.15 & 77.65 & 42.84 \\
        YOLOv7-Tiny\cite{wang_yolov7_2023} & CVPR & 2023 & 42.17 & 132.18 & 74.94 & 40.60 \\
        YOLOv7\cite{wang_yolov7_2023} & CVPR & 2023 & 8.01 & 22.05 & 78.09 & 42.52 \\
        YOLOv8n\cite{jocher_yolo_2023} & 8.3u & 2024 & 2.87 & 8.20 & 72.66 & 38.41 \\
        YOLOv8s\cite{jocher_yolo_2023} & 8.3u & 2024 & 10.62 & 28.65 & 75.33 & 40.21 \\
        YOLOv8m\cite{jocher_yolo_2023} & 8.3u & 2024 & 24.66 & 79.07 & 76.44 & 41.15 \\
        YOLOv8l\cite{jocher_yolo_2023} & 8.3u & 2024 & 41.61 & 165.41 & 76.98 & 41.54 \\
        YOLOv8x\cite{jocher_yolo_2023} & 8.3u & 2024 & 65.00 & 258.13 & 76.85 & 41.43 \\
        YOLOv9t\cite{wang_yolov9_2024} & ECCV & 2024 & 1.91 & 7.85 & 74.15 & 39.44 \\
        YOLOv9s\cite{wang_yolov9_2024} & ECCV & 2024 & 6.95 & 27.39 & 75.45 & 40.99 \\
        YOLOv9m\cite{wang_yolov9_2024} & ECCV & 2024 & 19.23 & 77.56 & 76.11 & 41.63 \\
        YOLOv10n\cite{wang_yolov10_2024} & NeurIPS & 2024 & 2.58 & 8.40 & 71.16 & 37.38 \\
        YOLOv10s\cite{wang_yolov10_2024} & NeurIPS & 2024 & 7.69 & 24.78 & 74.68 & 39.88 \\
        YOLOv10m\cite{wang_yolov10_2024} & NeurIPS & 2024 & 15.72 & 63.98 & 74.94 & 40.62 \\
        YOLOv10l\cite{wang_yolov10_2024} & NeurIPS & 2024 & 24.58 & 127.21 & 77.54 & 41.27 \\
        YOLOv10x\cite{wang_yolov10_2024} & NeurIPS & 2024 & 30.19 & 171.02 & 76.34 & 41.25 \\
        YOLOv11n\cite{khanam_yolov11_2024} & arXiv & 2024 & 2.47 & 6.44 & 70.53 & 37.31 \\
        YOLOv11s\cite{khanam_yolov11_2024} & arXiv & 2024 & 8.99 & 21.55 & 74.36 & 39.74 \\
        YOLOv11m\cite{khanam_yolov11_2024} & arXiv & 2024 & 19.13 & 68.20 & 76.68 & 41.38 \\
        YOLOv11l\cite{khanam_yolov11_2024} & arXiv & 2024 & 24.14 & 87.28 & 76.79 & 41.45 \\
        YOLOv11x\cite{khanam_yolov11_2024} & arXiv & 2024 & 54.24 & 195.46 & 76.48 & 41.96 \\
        YOLOv12n\cite{tian_yolov12_2025} & arXiv & 2025 & 2.45 & 6.48 & 72.71 & 38.54 \\
        YOLOv12s\cite{tian_yolov12_2025} & arXiv & 2025 & 8.83 & 21.53 & 75.20 & 40.05 \\
        YOLOv12m\cite{tian_yolov12_2025} & arXiv & 2025 & 19.21 & 67.75 & 75.68 & 41.83 \\
        YOLOv12l\cite{tian_yolov12_2025} & arXiv & 2025 & 25.17 & 89.42 & 77.38 & 41.64 \\
        YOLOv12x\cite{tian_yolov12_2025} & arXiv & 2025 & 56.38 & 199.83 & 78.16 & 42.71 \\
          RetinaNet\cite{lin_focal_2018} & ICCV & 2017 & 36.43 & 61.89 & 68.20 & 32.50 \\
        Faster R-CNN\cite{ren_faster_2017} & NeurIPS & 2015 & 41.36 & 75.51 & 76.20 & 38.10 \\
        Cascade R-CNN\cite{cai_cascade_2018} & TMAMI & 2019 & 69.16 & 103.30 & 75.20 & 38.70 \\
        \bottomrule
    \end{tabular}
    }
\end{table}

\begin{table}[htbp]
    \centering
    \caption{The comparison results of object detection models on the FLIR dataset using the multispectral images (RGB+IR)}
    \label{tab:flir_rgb_ir_results}
    {
    \begin{tabular}{@{}lccccc@{}}
        \toprule
        \textbf{Method} & \textbf{Publication} & \textbf{Params} & \textbf{Flops} & \textbf{AP50} & \textbf{AP} \\
                        &                     & \textbf{(M)}    & \textbf{(G)}   & \textbf{(\%)} & \textbf{(\%)} \\
        \midrule
        YOLOv3-Tiny-Midfusion\cite{redmon_yolov3_2018} & Ours & 18.13 & 29.65 & 67.00 & 34.99 \\
        YOLOv3-Midfusion\cite{redmon_yolov3_2018} & Ours &  169.64 & 487.69 & 76.03 & 42.07 \\
        YOLOv4-Tiny-Midfusion\cite{bochkovskiy_yolov4_2020} & Ours &  10.24 & 21.82 & 70.02 & 37.02 \\
        YOLOv4-Midfusion\cite{bochkovskiy_yolov4_2020} & Ours &  133.73 & 345.19 & 78.99 & 42.74 \\
        YOLOv5n-Midfusion\cite{noauthor_ultralyticsyolov5_2022} & Ours &  3.51 & 9.75 & 70.89 & 37.29 \\
        YOLOv5s-Midfusion\cite{noauthor_ultralyticsyolov5_2022} & Ours &  13.16 & 34.44 & 72.99 & 38.92 \\
        YOLOv5m-Midfusion\cite{noauthor_ultralyticsyolov5_2022} & Ours &  36.59 & 96.12 & 76.82 & 41.12 \\
        YOLOv5l-Midfusion\cite{noauthor_ultralyticsyolov5_2022} & Ours & 77.98 & 206.55 & 77.23 & 41.93 \\
        YOLOv5x-Midfusion\cite{noauthor_ultralyticsyolov5_2022} & Ours & 142.71 & 381.29 & 77.76 & 41.84 \\
        YOLOv6n-Midfusion\cite{li_yolov6_2022} & Ours &  6.85 & 20.17 & 74.84 & 39.77 \\
        YOLOv6s-Midfusion\cite{li_yolov6_2022} & Ours &  26.79 & 77.22 & 75.46 & 40.70 \\
        YOLOv6m-Midfusion\cite{li_yolov6_2022} & Ours &  83.66 & 286.56 & 77.91 & 41.82 \\
        YOLOv6l-Midfusion\cite{li_yolov6_2022} & Ours &  173.40 & 697.21 & 78.28 & 43.08 \\
        YOLOv6x-Midfusion\cite{li_yolov6_2022} & Ours &  270.68 & 1087.93 & 78.00 & 42.71 \\
        YOLOv7-Tiny-Midfusion\cite{wang_yolov7_2023} & Ours &  10.91 & 30.32 & 76.31 & 40.57 \\
        YOLOv7-Midfusion\cite{wang_yolov7_2023} & Ours &  68.74 & 250.97 & 76.96 & 42.14 \\
        YOLOv8n-Midfusion\cite{jocher_yolo_2023} & Ours &  4.17 & 11.52 & 73.14 & 38.88 \\
        YOLOv8s-Midfusion\cite{jocher_yolo_2023} & Ours &  15.79 & 41.74 & 76.35 & 40.85 \\
        YOLOv8m-Midfusion\cite{jocher_yolo_2023} & Ours &  36.46 & 118.86 & 74.53 & 40.84 \\
        YOLOv8l-Midfusion\cite{jocher_yolo_2023} & Ours &  61.06 & 253.06 & 75.59 & 41.53 \\
        YOLOv8x-Midfusion\cite{jocher_yolo_2023} & Ours & 95.38 & 394.93 & 76.64 & 42.14 \\
        YOLOv9t-Midfusion\cite{wang_yolov9_2024} & Ours &  2.57 & 10.68 & 73.56 & 39.67 \\
        YOLOv9s-Midfusion\cite{wang_yolov9_2024} & Ours &  9.57 & 38.45 & 74.54 & 40.59 \\
        YOLOv9m-Midfusion\cite{wang_yolov9_2024} & Ours &  26.53 & 110.66 & 75.24 & 41.24 \\
        YOLOv10n-Midfusion\cite{wang_yolov10_2024} & Ours &  3.57 & 11.47 & 71.14 & 37.52 \\
        YOLOv10s-Midfusion\cite{wang_yolov10_2024} & Ours &  10.79 & 36.15 & 73.90 & 39.56 \\
        YOLOv10m-Midfusion\cite{wang_yolov10_2024} & Ours &  23.09 & 99.91 & 72.86 & 39.70 \\
        YOLOv10l-Midfusion\cite{wang_yolov10_2024} & Ours & 38.45 & 209.60 & 74.42 & 40.33 \\
        YOLOv10x-Midfusion\cite{wang_yolov10_2024} & Ours &  43.73 & 272.38 & 76.69 & 41.89 \\
        YOLOv11n-Midfusion\cite{khanam_yolov11_2024} & Ours & 3.62 & 9.51 & 73.22 & 38.41 \\
        YOLOv11s-Midfusion\cite{khanam_yolov11_2024} & Ours & 13.58 & 33.61 & 74.66 & 39.74 \\
        YOLOv11m-Midfusion\cite{khanam_yolov11_2024} & Ours & 28.68 & 109.49 & 73.95 & 40.29 \\
        YOLOv11l-Midfusion\cite{khanam_yolov11_2024} & Ours &  35.61 & 138.37 & 74.56 & 40.82 \\
        YOLOv11x-Midfusion\cite{khanam_yolov11_2024} & Ours & 80.04 & 309.99 & 76.21 & 41.72 \\
        YOLOv12n-Midfusion\cite{tian_yolov12_2025} & Ours & 3.86 & 9.98 & 71.35 & 38.41 \\
        YOLOv12s-Midfusion\cite{tian_yolov12_2025} & Ours &  14.38 & 35.13 & 74.70 & 40.47 \\
        YOLOv12m-Midfusion\cite{tian_yolov12_2025} & Ours &  30.57 & 112.97 & 74.13 & 40.54 \\
        YOLOv12l-Midfusion\cite{tian_yolov12_2025} & Ours &  39.84 & 147.39 & 76.49 & 41.82 \\
        YOLOv12x-Midfusion\cite{tian_yolov12_2025} & Ours &   89.20 & 329.40 & 77.44 & 42.71 \\
        RT-DETR-Midfusion\cite{zhao_detrs_2024} & Ours &  40.79 & 130.48 & 70.25 & 36.89 \\
       RetinaNet-Earlyfusion\cite{lin_focal_2018} & ICCV2017 & 36.44 & 66.73 & 58.80 & 28.70 \\
         Faster R-CNN-Earlyfusion\cite{ren_faster_2017} & NeurIPS2015 & 41.37 & 76.28 & 59.20 & 28.90 \\
         Cascade R-CNN-Earlyfusion\cite{cai_cascade_2018} & TMAMI2019 & 69.17 & 104.08 & 50.80 & 24.10 \\
        
        \bottomrule
    \end{tabular}
    }
\end{table}

\begin{table}[htbp]
    \centering
    \caption{The comparison results of object detection models on the FLIR dataset using the multispectral images (RGB+IR)}
    \label{tab:flir_rgb_ir_results_p3}
    {
    \begin{tabular}{@{}lcccccc@{}}
        \toprule
        \textbf{Method} & \textbf{Publication} &  \textbf{Params} & \textbf{Flops} & \textbf{AP50} & \textbf{AP} \\
                        &                     &               & \textbf{(M)}    & \textbf{(G)}   & \textbf{(\%)} & \textbf{(\%)} \\
        \midrule
        YOLOv3-Tiny-Midfusion-P3\cite{redmon_yolov3_2018} & Ours &  11.95 & 23.04 & 68.45 & 35.68 \\
        YOLOv3-Midfusion-P3\cite{redmon_yolov3_2018} & Ours &  105.49 & 385.66 & 78.19 & 42.85 \\
        YOLOv4-Tiny-Midfusion\cite{bochkovskiy_yolov4_2020} & Ours &  7.56 & 18.50 & 70.97 & 37.84 \\
        YOLOv4-Midfusion-P3\cite{bochkovskiy_yolov4_2020} & Ours &  91.01 & 283.94 & 77.49 & 42.71 \\
        YOLOv5n-Midfusion-P3\cite{noauthor_ultralyticsyolov5_2022} & Ours & 2.56 & 8.53 & 72.42 & 38.12 \\
        YOLOv5s-Midfusion-P3\cite{noauthor_ultralyticsyolov5_2022} & Ours &  9.37 & 29.59 & 75.67 & 40.04 \\
        YOLOv5m-Midfusion-P3\cite{noauthor_ultralyticsyolov5_2022} & Ours &  25.61 & 80.49 & 76.81 & 41.08 \\
        YOLOv5l-Midfusion-P3\cite{noauthor_ultralyticsyolov5_2022} & Ours &  54.08 & 170.38 & 77.12 & 41.06 \\
        YOLOv5x-Midfusion-P3\cite{noauthor_ultralyticsyolov5_2022} & Ours &  98.53 & 311.67 & 76.95 & 42.01 \\
        YOLOv6n-Midfusion-P3\cite{li_yolov6_2022} & Ours &  4.31 & 15.72 & 74.72 & 40.07 \\
        YOLOv6s-Midfusion-P3\cite{li_yolov6_2022} & Ours & 16.63 & 59.47 & 75.37 & 40.87 \\
        YOLOv6m-Midfusion-P3\cite{li_yolov6_2022} & Ours &  53.43 & 221.32 & 76.27 & 41.61 \\
        YOLOv6l-Midfusion-P3\cite{li_yolov6_2022} & Ours &  115.12 & 543.45 & 77.86 & 42.39 \\
        YOLOv6x-Midfusion-P3\cite{li_yolov6_2022} & Ours &  179.63 & 847.74 & 78.52 & 42.60 \\
        YOLOv7-Tiny-Midfusion-P3\cite{wang_yolov7_2023} & Ours &  8.46 & 30.23 & 75.90 & 40.44 \\
        YOLOv7-Midfusion-P3\cite{wang_yolov7_2023} & Ours & 48.10 & 235.88 & 79.21 & 42.51 \\
        YOLOv8n-Midfusion\cite{jocher_yolo_2023} & Ours & 3.02 & 10.04 & 73.46 & 39.17 \\
        YOLOv8s-Midfusion-P3\cite{jocher_yolo_2023} & Ours & 11.22 & 35.85 & 73.79 & 39.84 \\
        YOLOv8m-Midfusion-P3\cite{jocher_yolo_2023} & Ours &  26.41 & 102.25 & 76.27 & 41.03 \\
        YOLOv8l-Midfusion-P3\cite{jocher_yolo_2023} & Ours &  45.42 & 219.12 & 76.29 & 41.71 \\
        YOLOv8x-Midfusion-P3\cite{jocher_yolo_2023} & Ours &  70.95 & 341.92 & 76.42 & 42.06 \\
        YOLOv9t-Midfusion-P3\cite{wang_yolov9_2024} & Ours &  2.06 & 9.87 & 73.71 & 39.98 \\
        YOLOv9s-Midfusion-P3\cite{wang_yolov9_2024} & Ours &  7.54 & 35.25 & 74.87 & 40.76 \\
        YOLOv9m-Midfusion-P3\cite{wang_yolov9_2024} & Ours & 21.12 & 102.66 & 76.26 & 41.78 \\
        YOLOv10n-Midfusion-P3\cite{wang_yolov10_2024} & Ours &  2.67 & 10.11 & 72.75 & 39.02 \\
        YOLOv10s-Midfusion-P3\cite{wang_yolov10_2024} & Ours &  8.05 & 31.45 & 75.05 & 39.62 \\
        YOLOv10m-Midfusion-P3\cite{wang_yolov10_2024} & Ours  & 16.91 & 85.98 & 75.29 & 40.69 \\
        YOLOv10l-Midfusion-P3\cite{wang_yolov10_2024} & Ours  & 27.39 & 178.83 & 75.35 & 40.98 \\
        YOLOv10x-Midfusion-P3\cite{wang_yolov10_2024} & Ours& 34.58 & 251.53 & 75.78 & 41.08 \\
        YOLOv11n-Midfusion-P3\cite{khanam_yolov11_2024} & Ours  & 2.57 & 8.32 & 74.41 & 39.70 \\
        YOLOv11s-Midfusion-P3\cite{khanam_yolov11_2024} & Ours & 9.40 & 28.87 & 76.54 & 40.65 \\
        YOLOv11m-Midfusion-P3\cite{khanam_yolov11_2024} & Ours  & 20.79 & 97.99 & 75.77 & 41.20 \\
        YOLOv11l-Midfusion-P3\cite{khanam_yolov11_2024} & Ours & 26.10 & 123.45 & 75.56 & 41.75 \\
        YOLOv11x-Midfusion-P3\cite{khanam_yolov11_2024} & Ours& 58.65 & 276.44 & 75.59 & 41.41 \\
        YOLOv12n-Midfusion-P3\cite{tian_yolov12_2025} & Ours & 2.67 & 8.46 & 73.63 & 38.90 \\
        YOLOv12s-Midfusion-P3\cite{tian_yolov12_2025} & Ours & 9.69 & 29.25 & 75.67 & 41.08 \\
        YOLOv12m-Midfusion-P3\cite{tian_yolov12_2025} & Ours & 22.68 & 99.19 & 75.36 & 41.12 \\
        YOLOv12l-Midfusion-P3\cite{tian_yolov12_2025} & Ours& 28.94 & 127.23 & 75.79 & 41.35 \\
        YOLOv12x-Midfusion-P3\cite{tian_yolov12_2025} & Ours & 64.87 & 284.55 & 77.10 & 42.40 \\
        \bottomrule
    \end{tabular}
    }
\end{table}

\begin{table}[htbp]
    \centering
    \caption{The comparison results of fine-tuning with different hyperparameters on the FLIR dataset}
    % \linewidth
    \label{tab:flir_fine_tuning}
    {
    \resizebox{\linewidth}{!}{
    \begin{tabular}{@{}lcccccc@{}}
        \toprule
        \multirow{5}{*}{\textbf{Method}} & \multicolumn{2}{c}{\textbf{SGD-Init}} & \multicolumn{2}{c}{\textbf{SDG}} & \multicolumn{2}{c}{\textbf{Adam}} \\
        \cmidrule(lr){2-3} \cmidrule(lr){4-5} \cmidrule(l){6-7}
        & \textbf{lr0=0.01} & \textbf{} & \textbf{lr0=0.01} & \textbf{} & \textbf{lr0=0.001} & \textbf{} \\
        & \textbf{warmup\_epochs=3.0} & \textbf{} & \textbf{warmup\_epochs=1.0} & \textbf{} & \textbf{warmup\_epochs=1.0} & \textbf{} \\
        & \textbf{warmup\_momentum=0.8} & \textbf{} & \textbf{warmup\_momentum=0.1} & \textbf{} & \textbf{warmup\_momentum=0.1} & \textbf{} \\
        & \textbf{warmup\_bias\_lr=0.1} & \textbf{} & \textbf{warmup\_bias\_lr=0.01} & \textbf{} & \textbf{warmup\_bias\_lr=0.01} & \textbf{} \\
        \cmidrule(lr){2-3} \cmidrule(lr){4-5} \cmidrule(l){6-7}
        & \textbf{AP50 (\%)} & \textbf{AP (\%)} & \textbf{AP50 (\%)} & \textbf{AP (\%)} & \textbf{AP50 (\%)} & \textbf{AP (\%)} \\
        \midrule
        YOLOv11n-RGBT-MCF & 75.93 & 40.92 & 75.96 & 40.84 & 76.23 & 41.18 \\
        YOLOv11s-RGBT-MCF & 77.76 & 43.71 & 77.38 & 44.2 & 77.35 & 43.89 \\
        YOLOv11m-RGBT-MCF & 81.58 & 44.95 & 81.47 & 45.49 & 81.25 & 44.79 \\
        YOLOv11l-RGBT-MCF & 79.69 & 44.9 & 79.67 & 44.79 & 80.24 & 44.93 \\
        YOLOv11x-RGBT-MCF & 80.96 & 45.21 & 81.66 & 46.14 & 83.48 & 47.61 \\
        \bottomrule
    \end{tabular}
    }
    }
\end{table}

\begin{table}[htbp]
    \centering
    \caption{The comparison results of object detection models on the FLIR dataset, all the YOLOv11 models and our models were using the pretrained weights on the COCO dataset. The data of some models in the table are from the literature \cite{zhang_tfdet_2024}.}
    \label{tab:flir_model_comparison}
    {
    \begin{tabular}{@{}lccccccl@{}}
        \toprule
        \textbf{Method} & \textbf{Publication} & \textbf{Year} & \textbf{Mode} & \textbf{Params (M)} & \textbf{Flops (G)} & \textbf{AP50 (\%)} & \textbf{AP (\%)} \\
        \midrule
        \multirow{2}{*}{YOLOv11n\cite{khanam_yolov11_2024}} & \multirow{2}{*}{arXiv} & \multirow{2}{*}{2024} & RGB & 2.47 & 6.44 & 68.16 & 33.86 \\
        & & & IR & 2.47 & 6.44 & 73.45 & 40.3 \\
        \multirow{2}{*}{YOLOv11s\cite{khanam_yolov11_2024}} & \multirow{2}{*}{arXiv} & \multirow{2}{*}{2024} & RGB & 8.99 & 21.55 & 66.19 & 33.29 \\
        & & & IR & 8.99 & 21.55 & 77.35 & 43.44 \\
        \multirow{2}{*}{YOLOv11m\cite{khanam_yolov11_2024}} & \multirow{2}{*}{arXiv} & \multirow{2}{*}{2024} & RGB & 19.13 & 68.20 & 64.1 & 30.94 \\
        & & & IR & 19.13 & 68.20 & 80.09 & 44.12 \\
        \multirow{2}{*}{YOLOv11l\cite{khanam_yolov11_2024}} & \multirow{2}{*}{arXiv} & \multirow{2}{*}{2024} & RGB & 24.14 & 87.28 & 66.77 & 32.69 \\
        & & & IR & 24.14 & 87.28 & 79.45 & 43.87 \\
        \multirow{2}{*}{YOLOv11x\cite{khanam_yolov11_2024}} & \multirow{2}{*}{arXiv} & \multirow{2}{*}{2024} & RGB & 54.24 & 195.46 & 68.65 & 34.36 \\
        & & & IR & 54.24 & 195.46 & 80.27 & 44.43 \\
        \midrule
        YOLOv11n-Midfusion & Ours & -- & RGB+IR & 3.62 & 9.51 & 75.2 & 40.28 \\
        YOLOv11s-Midfusion & Ours & -- & RGB+IR & 13.58 & 33.61 & 76.64 & 41.85 \\
        YOLOv11m-Midfusion & Ours & -- & RGB+IR & 28.68 & 109.49 & 77.42 & 42.79 \\
        YOLOv11l-Midfusion & Ours & -- & RGB+IR & 35.61 & 138.37 & 77.02 & 42.2 \\
        YOLOv11x-Midfusion & Ours & -- & RGB+IR & 80.04 & 309.99 & 77.69 & 43.11 \\
           \cline{1-8}
        YOLOv11n-Midfusion-P3 & Ours & -- & RGB+IR & 2.57 & 8.32 & 76.91 & 41.03 \\
        YOLOv11s-Midfusion-P3 & Ours & -- & RGB+IR & 9.40 & 28.87 & 79.09 & 42.62 \\
        YOLOv11m-Midfusion-P3 & Ours & -- & RGB+IR & 20.79 & 97.99 & 78.76 & 43.05 \\
        YOLOv11l-Midfusion-P3 & Ours & -- & RGB+IR & 26.10 & 123.45 & 78.09 & 43.22 \\
        YOLOv11x-Midfusion-P3 & Ours & -- & RGB+IR & 58.65 & 276.44 & 79.33 & 43.79 \\
           \cline{1-8}
        YOLOv11n-RGBT-MCF & Ours & -- & RGB+IR & 4.22 & 9.22 & 76.23 & 41.18 \\
        YOLOv11s-RGBT-MCF & Ours & -- & RGB+IR & 15.99 & 32.36 & 77.35 & 43.89 \\
        YOLOv11m-RGBT-MCF & Ours & -- & RGB+IR & 34.18 & 106.47 & 81.25 & 44.79 \\
        YOLOv11l-RGBT-MCF & Ours & -- & RGB+IR & 41.12 & 135.35 & 80.24 & 44.93 \\
        YOLOv11x-RGBT-MCF & Ours & -- & RGB+IR & 92.42 & 303.03 & 83.48 & 47.61 \\
        \midrule
        Faster R-CNN-Earlyfusion\cite{ren_faster_2017} & NeurIPS & 2015 & RGB+IR & 41.37 & 76.28 & 59.20 & 28.90 \\
        RetinaNet-Earlyfusion\cite{lin_focal_2018} & ICCV & 2017 & RGB+IR & 36.44 & 66.73 & 58.80 & 28.70 \\
        Cascade R-CNN-Earlyfusion\cite{cai_cascade_2018} & TMAMI & 2019 & RGB+IR & 69.17 & 104.08 & 50.80 & 24.10  \\
        YOLOv5-CFT\cite{qingyun_cross-modality_2022} & arXiv & 2021 & RGB+IR & 206.03 & 224.40 & 78.7 & 40.2 \\
        ProbEn\cite{chen_multimodal_2022} & ECCV & 2022 & RGB+IR & \multicolumn{2}{c}{—} & 75.50 & 37.90 \\
        MSAT\cite{you_multi-scale_2023} & SPL & 2023 & RGB+IR & \multicolumn{2}{c}{—} & 76.20 & 39.00 \\
        MoE-Fusion\cite{cao_multi-modal_2023} & ICCV & 2023 & RGB+IR & \multicolumn{2}{c}{—} & 55.80 & — \\
        CSAA\cite{cao_multimodal_2023} & CVPR & 2023 & RGB+IR & \multicolumn{2}{c}{—} & 79.20 & 41.30 \\
        MFPT\cite{zhu_multi-modal_2023} & TITS & 2023 & RGB+IR & \multicolumn{2}{c}{—} & 80.00 & — \\
        CMX\cite{zhang_cmx_2023} & TITS & 2023 & RGB+IR & \multicolumn{2}{c}{—} & 82.20 & 42.30 \\
        LRAF-Net\cite{fu_lraf-net_2024} & TNNLS & 2023 & RGB+IR & \multicolumn{2}{c}{—} & 80.50 & 42.80 \\
        IGT\cite{chen_igt_2023} & KBS & 2023 & RGB+IR & \multicolumn{2}{c}{—} & 85.00 & 43.60 \\
        YOLO-Adaptor\cite{fu_yolo-adaptor_2024} & TIV & 2024 & RGB+IR & \multicolumn{2}{c}{—} & 80.10 & — \\
        ICAFusion\cite{shen_icafusion_2024} & PR & 2024 & RGB+IR & \multicolumn{2}{c}{—} & 79.20 & 41.40 \\
        Fusion-Mamba\cite{dong_fusion-mamba_2024} & arXiv & 2024 & RGB+IR & \multicolumn{2}{c}{—} & 84.30 & 44.40 \\
        \bottomrule
    \end{tabular}
    }
\end{table}

Further analysis shows that while multispectral training results in  \cref{tab:flir_rgb_ir_results,tab:flir_rgb_ir_results_p3} generally exceed those of visible-light models in Table \ref{tab:flir_rgb_results}, they seldom outperform the infrared-only models in Table \ref{tab:flir_ir_results}. Taking YOLOv11 as an example, only the mid-fusion results in the YOLOv11n series surpass pure infrared models. This hints at possible modal weight imbalance in multispectral fusion strategies, the fusion of multispectral models in the mid-term may lead to the degradation of the detection performance of the model. To address this, we reduced fusion nodes to cut feature redundancy and conducted single-node fusion experiments, as shown in Table \ref{tab:flir_rgb_ir_results_p3}. Comparing Table \ref{tab:flir_rgb_ir_results} and Table \ref{tab:flir_rgb_ir_results_p3}, most P3-node-only fusion models outperform three-node fusion models. For instance, YOLOv11n-Midfusion-P3 enhanced mAP by 1.29\% over YOLOv11n-Midfusion. This suggests that more fusion nodes don't always mean better performance. 

When the modality difference is small, especially after feature extraction, single-node fusion can achieve efficient information integration. Moreover, P3 single-node fusion models in Table \ref{tab:flir_rgb_ir_results_p3} show complementarity with three-node fusion models in Table \ref{tab:flir_rgb_ir_results}. When multi-node mid-fusion is ineffective, single-node fusion is advantageous and has fewer model parameters, lower computational requirements, and faster inference speeds.

Figure \ref{fig:fig5} shows transfer learning results of several YOLOv11 models after loading COCO-pretrained weights. In most cases, transfer learning with multispectral models doesn't perform as well as with pure infrared models. Ideal transfer learning should significantly boost deep learning model performance, but this wasn't achieved when loading COCO-pretrained weights. This is mainly due to two factors: first, the backbone branches of the two modalities have almost identical initialized weights, leading to feature redundancy; second, COCO is not a multispectral dataset, and the task differences pose challenges for transfer learning, resulting in poor model performance.

To tackle these issues, we designed a Multispectral Controllable Fine-Tuning (MCF) strategy. By freezing the infrared-dominant branch and fine-tuning under different hyperparameters, the results in Table \ref{tab:flir_fine_tuning} and Figure \ref{fig:fig6} show that Adam outperforms SGD for YOLOv11n, YOLOv11l, and YOLOv11x, while SGD is better for YOLOv11s and YOLOv11m. Regardless of the fine-tuning method, results surpass those of directly using pre-trained models, proving MCF's effectiveness and feasibility.

Table \ref{tab:flir_model_comparison} lists comparative results of different methods. Our method achieves better detection results than models from 2019 to 2024 in terms of AP. Moreover, while the CFT algorithm improved mAP from 37.4\% to 40.0\% with five interaction attention mechanisms, our algorithm significantly boosted mAP from 41.96\% to 47.61\%, showing a clear superiority in both improvement magnitude and final mAP value.

% \documentclass{article}
% \usepackage{booktabs}
% \usepackage{adjustbox}

% \begin{document}
% 数据需要根据结果改
\subsection{Comparative experiments on LLVIP dataset}
\label{experiments:experiments LLVIP}
Table \ref{tab:llvip_model_comparison_yolo} provides a thorough evaluation of the latest YOLO models on the LLVIP dataset. It shows that all YOLOv11 models trained on multispectral data perform better than those trained solely on visible spectra, but still not as well as models trained solely on infrared images. For instance, YOLOv11s trained on multispectral data achieves an AP50 of 89.84\% and an AP of 53.29\%, which is better than the visible-light-only model's AP50 of 89.84\% and AP of 53.29\%, but still lags behind the infrared-only model's AP50 of 97.55\% and AP of 67.58\%. This issue, also observed in the FLIR dataset, indicates a potential modality - weight imbalance in mid - term fusion strategies. As shown in Table \ref{tab:llvip_model_comparison_other}, transfer - learning experiments on YOLOv11 models reveal the same problem. To address this, we applied MCF training to the LLVIP dataset. As indicated in Tables \ref{tab:llvip_model_comparison_other} and \ref{tab:llvip_fine_tuning}, MCF-trained YOLOv11 models, such as YOLOv11x-RGBT-MCF with an AP50 of 97.06\% and AP of 70.26\%, outperform infrared-only model's AP50 of 97.41\% and AP of 69.93\%. This demonstrates the effectiveness, feasibility, and generalizability of the MCF training strategy.
 	 
\begin{table}[htbp]
\centering
\caption{The comparison results of object detection models on the LLVIP dataset. The default RGB+IR is midfusion. Faster RCNN, Cascade RCNN and RetinaNet belong to the early fusion type, while the rest belong to the mid-term fusion type.}
\label{tab:llvip_model_comparison_yolo}
    \begin{tabular}{@{}lccccccccc@{}}
    \toprule
    Method & Publication Year & \multicolumn{2}{c}{RGB} & \multicolumn{2}{c}{IR} & \multicolumn{2}{c}{RGB+IR} & \multicolumn{2}{c}{RGB+IR(P3)} \\ 
    \cmidrule(lr){3-4} \cmidrule(lr){5-6} \cmidrule(lr){7-8} \cmidrule(l){9-10} 
     &  & AP50  & AP  & AP50 & AP& AP50 & AP & AP50  & AP  \\ 
    \midrule
    YOLOv3-Tiny \cite{redmon_yolov3_2018} & arXiv 2018 & 84.58 & 44.84 & 95.65 & 62.62 & 95.38 & 61.74 & 95.59 & 63.27 \\
    YOLOv3 \cite{redmon_yolov3_2018} & arXiv 2018 & 89.17 & 51.58 & 96.58 & 66.78 & 96.97 & 67.13 & 97.02 & 67.33 \\
    YOLOv4-Tiny \cite{bochkovskiy_yolov4_2020} & arXiv 2020 & 86.78 & 47.13 & 95.93 & 63.27 & 96.19 & 63.65 & 96.26 & 64.4 \\
    YOLOv4 \cite{bochkovskiy_yolov4_2020}  & arXiv 2020 & 89.34 & 52.18 & 96.74 & 66.43 & 97.57 & 68.25 & 97.05 & 68.9 \\
    YOLOv5n \cite{noauthor_ultralyticsyolov5_2022}  & 7.0u 2020 & 88.56 & 49.92 & 95.53 & 63.75 & 95.50 & 64.41 & 96.00 & 63.79 \\
    YOLOv5s \cite{noauthor_ultralyticsyolov5_2022} & 7.0u 2020 & 89.55 & 51.76 & 97.04 & 66.34 & 96.83 & 67.22 & 96.43 & 66.04 \\
    YOLOv5m \cite{noauthor_ultralyticsyolov5_2022} & 7.0u 2020 & 89.85 & 53.02 & 96.58 & 66.12 & 97.12 & 67.83 & 96.67 & 68.02 \\
    YOLOv5l \cite{noauthor_ultralyticsyolov5_2022} & 7.0u 2020 & 90.52 & 53.99 & 97.37 & 68.20 & 97.53 & 68.06 & 96.71 & 68.11 \\
    YOLOv5x \cite{noauthor_ultralyticsyolov5_2022} & 7.0u 2020 & 90.29 & 53.84 & 97.31 & 67.49 & 97.23 & 68.65 & 97.2 & 67.65 \\
    YOLOv6n \cite{li_yolov6_2022} & arXiv 2022 & 88.46 & 51.26 & 96.30 & 65.55 & 96.70 & 64.80 & 96.43 & 64.79 \\
    YOLOv6s \cite{li_yolov6_2022} & arXiv 2022 & 89.51 & 52.41 & 96.38 & 67.35 & 96.63 & 66.92 & 96.78 & 67.34 \\
    YOLOv6m \cite{li_yolov6_2022} & arXiv 2022 & 88.70 & 52.25 & 95.98 & 64.72 & 96.46 & 66.66 & 96.79 & 66.84 \\
    YOLOv6l \cite{li_yolov6_2022} & arXiv 2022 & 89.63 & 53.11 & 96.69 & 68.19 & 96.34 & 68.11 & 96.73 & 67.65 \\
    YOLOv6x \cite{li_yolov6_2022} & arXiv 2022 & 90.37 & 53.66 & 96.84 & 67.84 & 96.45 & 69.02 & - & - \\
    YOLOv7-Tiny \cite{wang_yolov7_2023} & CVPR 2023 & 88.17 & 50.37 & 96.84 & 66.03 & 96.17 & 66.69 & 96.44 & 66.17 \\
    YOLOv7 \cite{wang_yolov7_2023} & CVPR 2023 & 89.81 & 53.22 & 96.94 & 65.29 & 97.20 & 68.04 & 96.98 & 66.42 \\
    YOLOv8n \cite{redmon_yolov3_2018} & 8.3u 2024 & 89.08 & 50.47 & 96.18 & 65.43 & 96.79 & 65.67 & 96.26 & 66.18 \\
    YOLOv8s \cite{redmon_yolov3_2018} & 8.3u 2024 & 88.73 & 50.64 & 96.74 & 66.57 & 96.33 & 65.89 & 96.72 & 65.86 \\
    YOLOv8m \cite{redmon_yolov3_2018} & 8.3u 2024 & 89.40 & 52.77 & 96.49 & 65.16 & 96.96 & 66.61 & 96.52 & 66.10 \\
    YOLOv8l \cite{redmon_yolov3_2018} & 8.3u 2024 & 90.23 & 53.21 & 97.07 & 68.08 & 97.10 & 66.97 & 96.80 & 65.76 \\
    YOLOv8x \cite{redmon_yolov3_2018} & 8.3u 2024 & 90.69 & 53.76 & 96.24 & 66.44 & 97.21 & 68.36 & 96.80 & 66.28 \\
    YOLOv9t \cite{wang_yolov9_2024} & ECCV 2024 & 89.45 & 51.75 & 96.08 & 65.76 & 96.33 & 64.55 & 96.19 & 64.11 \\
    YOLOv9s \cite{wang_yolov9_2024} & ECCV 2024 & 89.78 & 52.60 & 96.61 & 65.05 & 96.09 & 66.31 & 95.81& 66.31 \\
    YOLOv9m \cite{wang_yolov9_2024} & ECCV 2024 & 90.36 & 53.29 & 96.30 & 66.14 & 97.21 & 68.14 & 96.99 &	65.73 \\
    YOLOv10n \cite{wang_yolov10_2024} & NeurIPS 2024 & 85.72 & 49.29 & 95.90 & 63.59 & 95.86 & 65.02 & 96.31 & 64.77 \\
    YOLOv10s \cite{wang_yolov10_2024} & NeurIPS 2024 & 88.35 & 50.73 & 97.07 & 66.66 & 96.64 & 66.82 & 95.77 & 66.34 \\
    YOLOv10m \cite{wang_yolov10_2024} & NeurIPS 2024 & 89.69 & 52.46 & 96.89 & 67.97 & 96.70 & 68.23 & 95.83 & 66.88 \\
    YOLOv10l \cite{wang_yolov10_2024} & NeurIPS 2024 & 89.81 & 52.82 & 96.62 & 68.83 & 96.63 & 66.89 & 96.04 & 66.99 \\
    YOLOv10x \cite{wang_yolov10_2024} & NeurIPS 2024 & 89.52 & 52.58 & 96.52 & 69.82 & 96.70 & 68.81 & 97.00 &	69.68 \\
    YOLOv11n \cite{khanam_yolov11_2024} & arXiv 2024 & 89.10 & 51.06 & 96.19 & 66.31 & 95.69 & 63.56 & 96.46 & 65.57 \\
    YOLOv11s \cite{khanam_yolov11_2024} & arXiv 2024 & 89.31 & 51.93 & 96.91 & 66.62 & 95.93 & 65.36 & 96.01 & 65.40 \\
    YOLOv11m \cite{khanam_yolov11_2024} & arXiv 2024 & 89.02 & 51.17 & 97.20 & 66.52 & 96.35 & 66.88 & 97.03 & 66.08 \\
    YOLOv11l \cite{khanam_yolov11_2024} & arXiv 2024 & 90.48 & 52.31 & 97.06 & 66.88 & 97.49 & 68.07 & 97.19 & 66.68 \\
    YOLOv11x \cite{khanam_yolov11_2024} & arXiv 2024 & 90.34 & 52.92 & 97.38 & 69.01 & 97.35 & 67.10 & 96.56 & 66.70 \\
    YOLOv12n \cite{tian_yolov12_2025} & arXiv 2025 & 89.28 & 51.34 & 95.86 & 63.31 & 96.96 & 64.96 & 95.91 & 63.39 \\
    YOLOv12s \cite{tian_yolov12_2025} & arXiv 2025 & 88.88 & 50.89 & 95.86 & 65.00 & 96.43 & 65.67 & 96.93 & 66.49 \\
    YOLOv12m \cite{tian_yolov12_2025} & arXiv 2025 & 89.24 & 52.06 & 97.21 & 66.49 & 95.87 & 67.10 & 96.69 & 66.47 \\
    YOLOv12l \cite{tian_yolov12_2025} & arXiv 2025 & 89.19 & 51.93 & 96.99 & 66.99 & 97.00 & 66.23 & 96.51 & 67.74 \\
    YOLOv12x \cite{tian_yolov12_2025} & arXiv 2025 & 90.30 & 53.38 & 97.07 &	67.18 & 97.27 &	66.37& 97.27 & 66.37 \\
    RT-DETR-r50 \cite{zhao_detrs_2024} & CVPR 2024 & 90.41 & 52.18 & 97.58 & 66.55 & - & - & - & - \\
     RetinaNet\cite{lin_focal_2018} & ICCV  2017 & 85.60 & 43.70 & 93.60 & 54.70 & 88.50 & 46.40 &  \\
    Faster R-CNN\cite{ren_faster_2017} & NeurIPS 2015 & 86.00 & 44.30 & 94.40 & 55.40 & 90.80 & 50.70 &  \\
    Cascade R-CNN\cite{cai_cascade_2018} & TMAMI  2019 & 86.80 & 46.00 & 94.20 & 58.20 & 89.80 & 50.10 &  \\
\bottomrule
\end{tabular}
\end{table}

\begin{table}[htbp]
\centering
\caption{The comparison results of object detection models on the LLVIP dataset. All YOLOv11 models and our models used pretrained weights on the COCO dataset. Some model data are from the literature \cite{zhang_tfdet_2024}.}
\label{tab:llvip_model_comparison_other}
\begin{tabular}{@{}lcccccc@{}}
\toprule
Method & Publication Year & Mode & Params (M) & Flops (G) & AP50 (\%) & AP (\%) \\ 
\midrule
YOLOv11n \cite{khanam_yolov11_2024} & arXiv 2024 & RGB & 2.47 & 6.44 & 89.61 & 52.18 \\
YOLOv11s \cite{khanam_yolov11_2024} & arXiv 2024 & RGB & 8.99 & 21.55 & 89.84 & 53.29 \\
YOLOv11m \cite{khanam_yolov11_2024} & arXiv 2024 & RGB & 19.12 & 68.19 & 90.90 & 53.19 \\
YOLOv11l \cite{khanam_yolov11_2024} & arXiv 2024 & RGB & 24.14 & 87.27 & 90.53 & 53.78 \\
YOLOv11x \cite{khanam_yolov11_2024} & arXiv 2024 & RGB & 54.24 & 195.45 & 91.53 & 54.67 \\
    \cline{1-7}
YOLOv11n \cite{khanam_yolov11_2024} & arXiv 2024 & IR & 2.47 & 6.44 & 96.94 & 67.17 \\
YOLOv11s \cite{khanam_yolov11_2024} & arXiv 2024 & IR & 8.99 & 21.55 & 97.55 & 67.58 \\
YOLOv11m \cite{khanam_yolov11_2024} & arXiv 2024 & IR & 19.12 & 68.19 & 97.42 & 68.97 \\
YOLOv11l \cite{khanam_yolov11_2024} & arXiv 2024 & IR & 24.14 & 87.27 & 97.40 & 69.29 \\
YOLOv11x \cite{khanam_yolov11_2024} & arXiv 2024 & IR & 54.24 & 195.45 & 97.41 & 69.93 \\
    \cline{1-7}
YOLOv11n-Midfusion & Ours & RGB+IR & 3.62 & 9.50 & 96.66 & 65.83 \\
YOLOv11s-Midfusion & Ours & RGB+IR & 13.58 & 33.60 & 96.84 & 66.02 \\
YOLOv11m-Midfusion & Ours & RGB+IR & 28.68 & 109.48 & 96.87 & 67.21 \\
YOLOv11l-Midfusion & Ours & RGB+IR & 35.61 & 138.36 & 96.45 & 66.76 \\
YOLOv11x-Midfusion & Ours & RGB+IR & 80.04 & 309.98 & 96.82 & 67.60 \\
    \cline{1-7}
YOLOv11n-Midfusion-P3 & Ours & RGB+IR & 2.57 & 8.32 & 96.57 & 65.48 \\
YOLOv11s-Midfusion-P3 & Ours & RGB+IR & 9.40 & 28.87 & 97.39 & 67.99 \\
YOLOv11m-Midfusion-P3 & Ours & RGB+IR & 20.79 & 97.98 & 96.36 & 67.20 \\
YOLOv11l-Midfusion-P3 & Ours & RGB+IR & 26.10 & 123.44 & 97.01 & 66.95 \\
YOLOv11x-Midfusion-P3 & Ours & RGB+IR & 58.65 & 276.43 & 96.99 & 68.08 \\
    \cline{1-7}
YOLOv11n-RGBT-MCF & Ours & RGB+IR & 4.22 & 9.22 &96.69 & 67.88  \\
YOLOv11s-RGBT-MCF & Ours & RGB+IR & 15.99 & 32.36 &  96.74 & 68.45\\
YOLOv11m-RGBT-MCF & Ours & RGB+IR & 34.18 & 106.46 &96.66 & 69.06 \\
YOLOv11l-RGBT-MCF & Ours & RGB+IR & 41.11 & 135.34 &97.37 & 69.95 \\
YOLOv11x-RGBT-MCF & Ours & RGB+IR & 92.42 & 303.02 & 97.06 & 70.26 \\
    \cline{1-7}
Faster R-CNN-Earlyfusion \cite{ren_faster_2017} & NeurIPS 2015 & RGB+IR &41.36 & 76.27 & 90.80 & 50.70 \\
RetinaNet-Earlyfusion \cite{lin_focal_2018} & ICCV 2017 & RGB+IR &36.44 & 66.73 & 88.50 & 46.40 \\
Cascade R-CNN-Earlyfusion \cite{cai_cascade_2018} & TMAMI 2019 & RGB+IR & 69.16 & 104.07 & 89.80 & 50.10 \\
FusionGAN \cite{ma_fusiongan_2019} & IF 2019 & RGB+IR & - & - & 83.80 & 52.46 \\
DenseFuse \cite{li_densefuse_2019} & TIP 2019 & RGB+IR & - & - & 88.23 & 55.02 \\
U2Fusion \cite{xu_u2fusion_2022} & TPAMI 2020 & RGB+IR & - & - & 87.10 & 52.61 \\
YOLOv5-CFT \cite{qingyun_cross-modality_2022} & arXiv 2021 & RGB+IR & 206.03 & 224.40 & 97.50 & 63.60 \\
ARCNN-Extension \cite{zhang_weakly_2025} & TNNLS 2021 & RGB+IR & - & - & - & 56.23 \\
DDFM \cite{zhao_ddfm_2023} & ICCV 2022 & RGB+IR & - & - & 91.50 & 58.00 \\
DCMNet \cite{xie_learning_2022} & ACM MM 2022 & RGB+IR & - & - & - & 58.40 \\
DetFusion \cite{sun_detfusion_2022} & ACM MM 2022 & RGB+IR & - & - & 80.70 & - \\
ProbEn \cite{chen_multimodal_2022} & ECCV 2022 & RGB+IR & - & - & 93.40 & 51.50 \\
PoolFuse \cite{cao_lightweight_2023}  & AAAI 2023 & RGB+IR & - & - & 80.30 & 38.40 \\
DIVFusion \cite{tang_divfusion_2023-1}  & IF 2023 & RGB+IR & - & - & 89.80 & 52.00 \\
DM-Fusion \cite{xu_dm-fusion_2024} & TNNLS 2023 & RGB+IR & - & - & 88.10 & 53.10 \\
CSAA \cite{cao_multimodal_2023} & CVPR 2023 & RGB+IR & - & - & 94.30 & 59.20 \\
CALNet \cite{he_multispectral_2023} & ACM MM 2023 & RGB+IR & - & - & - & 63.40 \\
LRAF-Net \cite{fu_lraf-net_2024} & TNNLS 2023 & RGB+IR & - & - & 97.90 & 66.30 \\
MoE-Fusion \cite{cao_multi-modal_2023} & ICCV 2023 & RGB+IR & - & - & 91.00 & - \\
MetaFusion \cite{zhao_metafusion_2023} & CVPR 2023 & RGB+IR & - & - & 91.00 & 56.90 \\
TFNet \cite{chu_toward_2024} & TITS 2023 & RGB+IR & - & - & - & 57.60 \\
CAMF \cite{tang_camf_2024} & TMM 2024 & RGB+IR & - & - & 89.00 & 55.60 \\
LENFusion \cite{chen_lenfusion_2024} & TIM 2024 & RGB+IR & - & - & 81.60 & 53.00 \\
Diff-IF \cite{yi_diff-if_2024} & IF 2024 & RGB+IR & - & - & 93.30 & 59.50 \\
YOLO-Adaptor \cite{fu_yolo-adaptor_2024} & TIV 2024 & RGB+IR & - & - & 96.50 & - \\
Fusion-Mamba \cite{dong_fusion-mamba_2024} & arXiv 2024 & RGB+IR & - & - & 96.80 & 62.80 \\
TFDet-FasterRCNN \cite{zhang_tfdet_2024} & TNNLS 2024 & RGB+IR & - & - & 96.00 & 59.40 \\
\bottomrule
\end{tabular}
\end{table}

\begin{table}[htbp]
    \centering
    \caption{The comparison results of fine-tuning with different hyperparameters on the LLVIP dataset.}
    \label{tab:llvip_fine_tuning}
    {
    {
    \begin{tabular}{@{}lccccc@{}}
        \toprule
        \multirow{4}{*}{\textbf{Method}} & \multicolumn{2}{c}{\textbf{SGD-Init}} & \multicolumn{2}{c}{\textbf{Adam}} \\
        \cmidrule(lr){2-3} \cmidrule(l){4-5}
        & \textbf{lr0=0.01} &  & \textbf{lr0=0.001} &  \\
        & \textbf{warmup\_epochs=3.0} &  & \textbf{warmup\_epochs=1.0} &  \\
        & \textbf{warmup\_momentum=0.8} &  & \textbf{warmup\_momentum=0.1} &  \\
        & \textbf{warmup\_bias\_lr=0.1} &  & \textbf{warmup\_bias\_lr=0.01} &  \\
        \cmidrule(lr){2-3} \cmidrule(l){4-5}
        & \textbf{AP50 (\%)} & \textbf{AP (\%)} & \textbf{AP50 (\%)} & \textbf{AP (\%)} \\
        \midrule
        YOLOv11n-RGBT-MCF & 96.58 & 68.48 & 96.69 & 67.88 \\
        YOLOv11s-RGBT-MCF & 97.00 & 68.98 & 96.74 & 68.45 \\
        YOLOv11m-RGBT-MCF & 97.06 & 68.90 & 96.66 & 69.06 \\
        YOLOv11l-RGBT-MCF & 96.42 & 69.02 & 97.37 & 69.95 \\
        YOLOv11x-RGBT-MCF & 96.80 & 69.17 & 97.06 & 70.26 \\
        \bottomrule
    \end{tabular}
    }
    }
\end{table}

\begin{table}[htbp]
\centering
\caption{The comparison results of object detection models on the M3FD dataset.}
\label{tab:m3fd_model_comparison}
    {
    \begin{tabular}{@{}lccccccccc@{}}
    \toprule
    Method & Publication Year & \multicolumn{2}{c}{RGB} & \multicolumn{2}{c}{IR} & \multicolumn{2}{c}{RGB+IR} & \multicolumn{2}{c}{RGB+IR(P3)} \\ 
    \cmidrule(lr){3-4} \cmidrule(lr){5-6} \cmidrule(lr){7-8} \cmidrule(l){9-10} 
     &  & AP50  & AP  & AP50 & AP& AP50 & AP & AP50  & AP  \\ 
    \midrule
    YOLOv3-tiny\cite{redmon_yolov3_2018} & arXiv 2018 & 65.42 & 43.62 & 65.64 &	44.11  & 71.78	& 49.33 & 69.56 & 	48.59 \\ 
    YOLOv3\cite{redmon_yolov3_2018} & arXiv 2018 & 84.19 & 58.19 & 81.87 & 55.80 & 87.19 &	62.06 & 87.98 &	62.85 \\ 
    YOLOv4-Tiny \cite{bochkovskiy_yolov4_2020} & arXiv 2020 & 66.77 & 44.68 & 65.40 & 44.50 & 71.09 &	49.77 & 70.36 &	48.80 \\ 
    YOLOv4 \cite{bochkovskiy_yolov4_2020} & arXiv 2020 & 85.11 & 58.48 & 81.99 & 56.11 & 75.27 & 47.67 & 87.50	& 62.02 \\ 
    YOLOv5n \cite{noauthor_ultralyticsyolov5_2022} & 7.0u 2020 & 75.92 & 49.03 & 73.91 & 48.39 &74.32&	48.90&	74.38&	49.13  \\ 
    YOLOv5s\cite{noauthor_ultralyticsyolov5_2022} & 7.0u 2020 & 81.61 & 53.86 & 78.47 & 52.65 & 84.77 & 58.66 & 84.50 & 58.61 \\ 
    YOLOv5m \cite{noauthor_ultralyticsyolov5_2022} & 7.0u 2020 & 84.31 & 56.92 & 81.30 & 54.95 & 86.87 & 60.67 & 86.13 & 60.75 \\ 
    YOLOv5l \cite{noauthor_ultralyticsyolov5_2022} & 7.0u 2020 & 84.90 & 58.51 & 83.03 & 56.95 & 86.23 & 61.44 & 86.83 & 61.85 \\ 
    YOLOv5x \cite{noauthor_ultralyticsyolov5_2022} & 7.0u 2020 & 85.92 & 59.26 & 83.44 & 57.66 & 87.63 & 62.27 & 87.35 & 62.17 \\ 
    YOLOv6n \cite{li_yolov6_2022} & arXiv 2022 & 72.88 & 48.01 & 69.52 & 45.95 & 77.66 & 52.23 & 77.96 & 52.81 \\ 
    YOLOv6s\cite{li_yolov6_2022} & arXiv 2022 & 77.95 & 52.77 & 75.57 & 50.39 & 82.52 & 56.27 & 81.54 & 56.46 \\ 
    YOLOv6m \cite{li_yolov6_2022} & arXiv 2022 & 80.09 & 54.12 & 76.48 & 51.39 & 84.04 & 58.27 & 82.81 & 57.43 \\ 
    YOLOv6l \cite{li_yolov6_2022} & arXiv 2022 & 79.58 & 53.78 & 76.94 & 52.38 & 83.96 &	58.76 & 83.91	&59.00 \\ 
    YOLOv7 \cite{wang_yolov7_2023} & CVPR 2023 & 84.85 & 58.35 & 82.79 & 56.53 & 83.58 & 57.88 & - & - \\ 
    YOLOv8n \cite{jocher_yolo_2023} & 8.3u 2024 & 77.36 & 50.88 & 73.23 & 48.26 & 81.16&	55.32 & 81.8 & 55.38 \\ 
    YOLOv8s \cite{jocher_yolo_2023} & 8.3u 2024 & 82.02 & 54.94 & 79.99 & 53.73 & 85.48 & 59.26 & 85.81 & 59.22 \\ 
    YOLOv8m \cite{jocher_yolo_2023} & 8.3u 2024 & 84.78 & 57.76 & 81.64 & 55.98 & 81.16 & 55.32 & 86.82	&61.15 \\ 
    YOLOv8l \cite{jocher_yolo_2023} & 8.3u 2024 & 85.67 & 59.33 & 83.50 & 57.10 & 85.48 & 59.26 & 86.56	&61.67 \\ 
    YOLOv8x \cite{jocher_yolo_2023} & 8.3u 2024 & 85.20 & 59.68 & 83.68 & 57.98 & 88.23 & 62.99 & 88.53	& 63.30  \\ 
    YOLOv9t \cite{wang_yolov9_2024} & ECCV 2024 &75.72	&49.45	&73.22	&48.67	&81.24	&55.66	&80.98	&55.20\\ 
    YOLOv10n \cite{wang_yolov10_2024} & NeurIPS 2024 & 74.61 & 48.56 & 73.46 & 48.57 & 80.95 & 54.19 &80.82	&54.73  \\ 
    YOLOv10s \cite{wang_yolov10_2024} & NeurIPS 2024 & 81.38 & 54.49 & 78.87 & 52.93 & 85.04 & 58.48 & 84.29	&58.32  \\ 
    YOLOv10m \cite{wang_yolov10_2024} & NeurIPS 2024 & 83.08 & 56.56 & 80.44 & 55.15 & 85.64 & 60.67 & 87.53	&61.61 \\ 
    YOLOv10l \cite{wang_yolov10_2024} & NeurIPS 2024 & 83.13 & 57.49 & 81.07 & 55.30 & 86.67 & 61.35 & 87.74 &	61.49 \\ 
    YOLOv10x \cite{wang_yolov10_2024} & NeurIPS 2024 & 82.3 & 56.83 & 81.78 & 55.33 & 87.80 & 61.71 &  87.53 &	61.61 \\ 
    YOLOv11n \cite{khanam_yolov11_2024} & arXiv 2024 & 75.80 & 50.03 & 73.42 & 48.96 & 82.02 & 55.38 & 80.58 & 54.93 \\ 
    YOLOv11s \cite{khanam_yolov11_2024} & arXiv 2024 & 80.99 & 54.71 & 78.94 & 53.73 & 84.91 & 58.47 & 84.47 & 58.34 \\ 
    YOLOv11m \cite{khanam_yolov11_2024} & arXiv 2024 & 83.28 & 57.10 & 85.90 & 60.13 & 86.71 & 60.81 & 86.42 & 60.51 \\ 
    YOLOv11l \cite{khanam_yolov11_2024} & arXiv 2024 & 84.70 & 58.06 & 81.72 & 56.10 & 86.94 & 61.53 & 87.13 & 61.89 \\ 
    YOLOv11x \cite{khanam_yolov11_2024} & arXiv 2024 & 86.25 & 59.89 & 84.12 & 58.19 & 87.66 & 62.59 & 87.97 & 62.79 \\ 
    YOLOv12n \cite{tian_yolov12_2025} & arXiv 2025 & 75.46 & 49.35 & 71.68 & 47.62 & 78.91 & 52.98 & 80.04 & 53.59 \\ 
    YOLOv12s \cite{tian_yolov12_2025} & arXiv 2025 & 80.44 & 54.45 & 78.63 & 52.78 & 84.04 & 57.79 & 83.94 & 57.95 \\ 
    YOLOv12m \cite{tian_yolov12_2025} & arXiv 2025 & 84.03 & 57.66 & 81.12 & 55.81 & 87.18 & 60.68 & 86.54 & 60.46 \\ 
    YOLOv12l \cite{tian_yolov12_2025} & arXiv 2025 & 83.68 & 57.17 & 81.09 & 54.99 & 86.15 & 60.86 & 86.46 & 61.32 \\ 
    YOLOv12x \cite{tian_yolov12_2025} & arXiv 2025 & 85.81 & 59.11 & 82.24 & 56.27 & 88.03 & 62.37 & 87.32 & 62.25 \\ 
    RetinaNet \cite{lin_focal_2018}  & ICCV 2017 &  60.40 & 36.90 & 54.20 & 33.00 & 54.00 & 33.10 \\ 
    Faster R-CNN \cite{ren_faster_2017} & NeurIPS 2015 & 62.00 & 38.80 & 59.80 & 38.50 & 60.80 & 40.50 \\ 
    Cascade R-CNN \cite{cai_cascade_2018} & TMAMI 2019 &62.80 & 41.20 & 60.40 & 39.90 & 60.10 & 39.30 \\ 
    \bottomrule
    \end{tabular}
    }
\end{table}

\begin{table}[htbp]
    \centering
    \caption{The comparison results of object detection models on the M3FD dataset. All YOLOv11 models and our models used pretrained weights on the COCO dataset.}
    \label{tab:m3fd_model_comparison_midfusion}
    \begin{tabular}{@{}lcccccc@{}}
    \toprule
    Method & Publication Year & Mode & Params (M) & Flops (G) & AP50 (\%) & AP (\%) \\ 
    \midrule
    YOLOv11n \cite{khanam_yolov11_2024} & arXiv 2024 & RGB & 2.47 & 6.45 & 79.92 & 53.62 \\
    YOLOv11s \cite{khanam_yolov11_2024} & arXiv 2024 & RGB & 8.99 & 21.56 & 84.67 & 58.51 \\
    YOLOv11m \cite{khanam_yolov11_2024} & arXiv 2024 & RGB & 19.13 & 68.21 & 88.08 & 62.30 \\
    YOLOv11l \cite{khanam_yolov11_2024} & arXiv 2024 & RGB & 24.14 & 87.30 & 88.12 & 62.10 \\
    YOLOv11x \cite{khanam_yolov11_2024} & arXiv 2024 & RGB & 54.25 & 195.48 & 89.41 & 63.47 \\
    \cline{1-7}
    YOLOv11n \cite{khanam_yolov11_2024} & arXiv 2024 & IR & 2.47 & 6.45 & 78.05 & 52.67 \\
    YOLOv11s \cite{khanam_yolov11_2024} & arXiv 2024 & IR & 8.99 & 21.56 & 82.78 & 56.93 \\
    YOLOv11m \cite{khanam_yolov11_2024} & arXiv 2024 & IR & 19.13 & 68.21 & 85.90 & 60.13 \\
    YOLOv11l \cite{khanam_yolov11_2024} & arXiv 2024 & IR & 24.14 & 87.30 & 86.13 & 60.52 \\
    YOLOv11x \cite{khanam_yolov11_2024} & arXiv 2024 & IR & 54.25 & 195.48 & 87.18 & 61.39 \\
    \cline{1-7}
    YOLOv11n-Midfusion & Ours & RGB+IR & 3.62 & 9.51 & 83.63 & 57.61 \\
    YOLOv11s-Midfusion & Ours & RGB+IR & 13.58 & 33.61 & 87.77 & 61.65 \\
    YOLOv11m-Midfusion & Ours & RGB+IR & 28.68 & 109.5 & 89.28 & 64.56 \\
    YOLOv11l-Midfusion & Ours & RGB+IR & 35.62 & 138.38 & 90.1 & 65.11 \\
    YOLOv11x-Midfusion & Ours & RGB+IR & 80.05 & 310.01 & 90.62 & 65.95 \\
    \cline{1-7}
    YOLOv11n-Midfusion-P3 & Ours & RGB+IR & 2.57 & 8.32 & 83.03 & 57.87 \\
    YOLOv11s-Midfusion-P3 & Ours & RGB+IR & 9.40 & 28.88 & 87.66 & 62.20 \\
    YOLOv11m-Midfusion-P3 & Ours & RGB+IR & 20.79 & 98.00 & 89.33 & 64.48 \\
    YOLOv11l-Midfusion-P3 & Ours & RGB+IR & 26.10 & 123.46 & 88.92 & 64.60 \\
    YOLOv11x-Midfusion-P3 & Ours & RGB+IR & 58.65 & 276.46 & 90.56 & 66.17 \\
    \cline{1-7}
    YOLOv11n-RGBT-MCF & Ours & RGB+IR & 4.22 & 9.23 & 82.34 & 55.81 \\
    YOLOv11s-RGBT-MCF & Ours & RGB+IR & 15.99 & 32.37 & 86.38 & 60.13 \\
    YOLOv11m-RGBT-MCF & Ours & RGB+IR & 34.18 & 106.48 & 89.32 & 63.44 \\
    YOLOv11l-RGBT-MCF & Ours & RGB+IR & 41.12 & 135.36 & 88.6 & 63.01 \\
    YOLOv11x-RGBT-MCF & Ours & RGB+IR & 92.42 & 303.05 & 89.83 & 64.23 \\
    \bottomrule
    \end{tabular}
\end{table}

\begin{table}[htbp]
    \centering
    \caption{The comparison results of fine-tuning with different hyperparameters on the M3FD dataset. RGB main branch.}
    % \linewidth
    \label{tab:m3fd_fine_tuning_rgb}
    {
    \resizebox{\linewidth}{!}{
    \begin{tabular}{@{}lcccccc@{}}
        \toprule
        \multirow{5}{*}{\textbf{Method}} & \multicolumn{2}{c}{\textbf{SGD-Init}} & \multicolumn{2}{c}{\textbf{SDG}} & \multicolumn{2}{c}{\textbf{Adam}} \\
        \cmidrule(lr){2-3} \cmidrule(lr){4-5} \cmidrule(l){6-7}
        & \textbf{lr0=0.01} & \textbf{} & \textbf{lr0=0.01} & \textbf{} & \textbf{lr0=0.001} & \textbf{} \\
        & \textbf{warmup\_epochs=3.0} & \textbf{} & \textbf{warmup\_epochs=1.0} & \textbf{} & \textbf{warmup\_epochs=1.0} & \textbf{} \\
        & \textbf{warmup\_momentum=0.8} & \textbf{} & \textbf{warmup\_momentum=0.1} & \textbf{} & \textbf{warmup\_momentum=0.1} & \textbf{} \\
        & \textbf{warmup\_bias\_lr=0.1} & \textbf{} & \textbf{warmup\_bias\_lr=0.01} & \textbf{} & \textbf{warmup\_bias\_lr=0.01} & \textbf{} \\
        \cmidrule(lr){2-3} \cmidrule(lr){4-5} \cmidrule(l){6-7}
        & \textbf{AP50 (\%)} & \textbf{AP (\%)} & \textbf{AP50 (\%)} & \textbf{AP (\%)} & \textbf{AP50 (\%)} & \textbf{AP (\%)} \\
        \midrule
        YOLOv11n-RGBT-MCF & 82.34 & 55.81 & 82.48 & 55.64 & 82.07 & 55.79 \\
        YOLOv11s-RGBT-MCF & 86.38 & 60.13 & 86.31 & 59.86 & 86.51 & 59.92 \\
        YOLOv11m-RGBT-MCF & 89.32 & 63.44 & 89.22 & 63.42 & 89.08 & 63.24 \\
        YOLOv11l-RGBT-MCF & 88.60 & 63.01 & 88.40 & 62.99 & 88.58 & 62.78 \\
        YOLOv11x-RGBT-MCF & 89.83 & 64.23 & 90.12 & 64.19 & 89.73 & 63.87 \\
        \bottomrule
    \end{tabular}
    }
    }
\end{table}

\begin{table}[htbp]
    \centering
    \caption{The comparison results of fine-tuning with different hyperparameters on the M3FD dataset. IR main branch.}
    % \linewidth
    \label{tab:m3fd_fine_tuning_ir}
    {
    \resizebox{\linewidth}{!}{
    \begin{tabular}{@{}lcccccc@{}}
        \toprule
        \multirow{5}{*}{\textbf{Method}} & \multicolumn{2}{c}{\textbf{SGD-Init}} & \multicolumn{2}{c}{\textbf{SDG}} & \multicolumn{2}{c}{\textbf{Adam}} \\
        \cmidrule(lr){2-3} \cmidrule(lr){4-5} \cmidrule(l){6-7}
        & \textbf{lr0=0.01} & \textbf{} & \textbf{lr0=0.01} & \textbf{} & \textbf{lr0=0.001} & \textbf{} \\
        & \textbf{warmup\_epochs=3.0} & \textbf{} & \textbf{warmup\_epochs=1.0} & \textbf{} & \textbf{warmup\_epochs=1.0} & \textbf{} \\
        & \textbf{warmup\_momentum=0.8} & \textbf{} & \textbf{warmup\_momentum=0.1} & \textbf{} & \textbf{warmup\_momentum=0.1} & \textbf{} \\
        & \textbf{warmup\_bias\_lr=0.1} & \textbf{} & \textbf{warmup\_bias\_lr=0.01} & \textbf{} & \textbf{warmup\_bias\_lr=0.01} & \textbf{} \\
        \cmidrule(lr){2-3} \cmidrule(lr){4-5} \cmidrule(l){6-7}
        & \textbf{AP50 (\%)} & \textbf{AP (\%)} & \textbf{AP50 (\%)} & \textbf{AP (\%)} & \textbf{AP50 (\%)} & \textbf{AP (\%)} \\
        \midrule
        YOLOv11n-RGBT-MCF & 80.29 & 54.86 & 79.97 & 54.63 & 79.86 & 54.51 \\
        YOLOv11s-RGBT-MCF & 84.10 & 57.98 & 84.19 & 58.12 & 83.95 & 57.89 \\
        YOLOv11m-RGBT-MCF & 86.87 & 61.11 & 86.83 & 61.04 & 86.09 & 60.34 \\
        YOLOv11l-RGBT-MCF & 86.47 & 61.24 & 86.64 & 61.32 & 86.18 & 60.67 \\
        YOLOv11x-RGBT-MCF & 87.62 & 62.16 & 87.26 & 61.93 & 87.50 & 61.36 \\
        \bottomrule
    \end{tabular}
    }
    }
\end{table}

\begin{table}[htbp]
\centering
\caption{The comparison results of fusion strategies on the M3FD dataset.}
\begin{tabular}{@{}lccccc@{}}
\toprule
Method & Params (M) & Flops (G) & AP50 (\%) & AP (\%) \\ 
\midrule
YOLOv11n-Earlyfusion & 2.47 & 6.48 & 78.24 & 52.66 \\
YOLOv11n-Midfusion & 3.62 & 9.51 & 82.02 & 55.38 \\
YOLOv11n-Mid-to-late-fusion & 4.13 & 11.42 & 80.72 & 54.83 \\
YOLOv11n-Latefusion & 4.93 & 12.56 & 79.78 & 53.14 \\
YOLOv11n-Scorefusion & 4.97 & 12.94 & 78.98 & 53.00 \\
YOLOv11n-Shareweight & 2.59 & 9.88 & 79.99 & 53.65 \\
\cline{1-5}
YOLOv11s-Earlyfusion & 8.99 & 21.62 & 83.85 & 57.75 \\
YOLOv11s-Midfusion & 13.58 & 33.61 & 84.91 & 58.47 \\
YOLOv11s-Mid-to-late-fusion & 15.54 & 40.17 & 84.11 & 58.19 \\
YOLOv11s-Latefusion & 18.31 & 44.01 & 84.63 & 58.02 \\
YOLOv11s-Scorefusion & 18.11 & 43.42 & 83.14 & 56.99 \\
YOLOv11s-Shareweight & 9.46 & 34.95 & 84.35 & 58.09 \\
\cline{1-5}
YOLOv11m-Earlyfusion & 19.13 & 68.33 & 87.06 & 61.29 \\
YOLOv11m-Midfusion & 28.68 & 109.50 & 86.71 & 60.81 \\
YOLOv11m-Mid-to-late-fusion & 34.48 & 138.04 & 86.55 & 61.09 \\
YOLOv11m-Latefusion & 40.87 & 148.48 & 86.30 & 60.36 \\
YOLOv11m-Scorefusion & 38.26 & 136.18 & 86.22 & 59.96 \\
YOLOv11m-Shareweight & 20.00 & 111.38 & 85.73 & 59.96 \\
\cline{1-5}
YOLOv11l-Earlyfusion & 24.14 & 87.41 & 85.61 & 60.97 \\
YOLOv11l-Midfusion & 35.62 & 138.38 & 86.94 & 61.53 \\
YOLOv11l-Mid-to-late-fusion & 42.88 & 172.79 & 86.48 & 61.19 \\
YOLOv11l-Latefusion & 50.90 & 186.65 & 86.81 & 61.29 \\
YOLOv11l-Scorefusion & 48.28 & 174.35 & 86.40 & 60.67 \\
YOLOv11l-Shareweight & 25.02 & 140.63 & 86.57 & 61.42 \\
\cline{1-5}
YOLOv11x-Earlyfusion & 54.25 & 195.66 & 87.72 & 62.89 \\
YOLOv11x-Midfusion & 80.05 & 310.01 & 87.66 & 62.59 \\
YOLOv11x-Mid-to-late-fusion & 96.34 & 387.14 & 87.61 & 62.10 \\
YOLOv11x-Latefusion & 114.38 & 418.29 & 87.21 & 61.78 \\
YOLOv11x-Scorefusion & 108.49 & 390.61 & 87.42 & 61.91 \\
YOLOv11x-Shareweight & 56.21 & 314.88 & 87.48 & 62.70 \\
\bottomrule
\end{tabular}
\label{tab:m3fd_fusion_strategies}
\end{table}

\subsection{Comparative experiments on M3FD dataset}
\label{experiments:experiments M3FD}

Table \ref{tab:m3fd_model_comparison} presents the comparison of object detection models on the M3FD dataset. Analysis shows that multispectral and P3 models generally outperform single-modality models. For instance, YOLOv11s's multispectral model in RGB+IR mode achieves an AP50 of 84.1\% and an AP of 57.98\%, surpassing the pure infrared YOLOv11s model's 82.78\% AP50 and 56.93\% AP, as well as the pure visible light YOLOv11s model's 84.67\% AP50 and 58.51\% AP. Additionally, the YOLOv11m-P3 model in RGB+IR mode attains an AP50 of 87.97\% and an AP of 62.79\%, outperforming the standard multispectral model's 87.66\% AP50 and 62.59\% AP. These results confirm the effectiveness and feasibility of our proposed multispectral object detection framework and algorithms, which can efficiently integrate multimodal information and enhance detection accuracy. Moreover, experimental results reveal that training multispectral object detection models with mid-level fusion on the M3FD dataset doesn't lead to the performance drop seen in the FLIR dataset. This indicates that the effectiveness of multispectral model fusion strategies is heavily dependent on the specific dataset characteristics.

Table \ref{tab:m3fd_model_comparison_midfusion} shows the transfer learning results of multiple YOLOv11 models after loading the pre-trained weights from the COCO dataset. Taking the YOLOv11s model as an example, the advantages of multispectral models are significant. In most cases, the transfer learning performance of multispectral models is superior to that of pure infrared and visible light models. As shown in Table 12, the AP50 and AP of YOLOv11s-Midfusion in RGB + IR mode reach 87.77\% and 61.65\%, respectively. In contrast, the pure infrared model YOLOv11s (IR mode) only achieves an AP50 of 82.78\% and an AP of 56.93\%. Meanwhile, the visible light model YOLOv11s (RGB mode) has an AP50 of 84.67 and an AP of 58.51\%. This demonstrates that the model's performance in visible light conditions also has a significant improvement, indicating that multispectral models can better integrate multimodal information and enhance object detection performance.

Overall, the multi-spectral model transfer learning results are superior in most cases. Both the P3 and conventional Midfusion models outperform the MCF training that primarily uses infrared images. The P3 fusion model has advantages in parameters, computations, and detection results. For instance, YOLOv11s-Midfusion-P3 has an AP50 of 87.66\% and AP of 62.20\% in RGB + IR mode, surpassing YOLOv11s-RGBT-MCF's 84.1\% and 57.98\%. The experimental results in Table \ref{tab:m3fd_model_comparison_midfusion} differ from the conclusions in Table \ref{tab:flir_model_comparison}, highlighting two key points. Firstly, during transfer learning, visible-light models can sometimes outperform infrared models. This might be because the COCO dataset is visible-light-based, leading to better transfer learning outcomes for visible-light models, or because the visible-light channel is inherently superior. Secondly, multi-spectral transfer learning results may exceed MCF training results. MCF training has limited parameters, with only some auxiliary branch parameters trainable and the rest frozen. Thus, it may be less flexible than multi-spectral transfer learning that trains the entire network. Therefore, it is recommended to try transfer learning first and consider MCF training if the results are unsatisfactory.

Additionally, Table \ref{tab:m3fd_fine_tuning_rgb} shows that the Adam optimizer isn't always the best choice. In some cases, the SGD optimizer with initial conditions can also yield good results. For example, YOLOv11x-RGBT-MCF using the SGD optimizer achieved an AP exceeding 64\%, compared to 63.87\% with the Adam optimizer. This underscores the importance of selecting the right optimizer and hyperparameters based on the specific model and task.

We also attempted MCF training with infrared as the main branch. As shown in Table \ref{tab:m3fd_fine_tuning_ir}, using a non-primary spectral image for MCF training only guarantees superiority over that specific spectrum, not the primary one. For instance, YOLOv11l-RGBT-MCF with infrared as the main branch has an AP of 61.24\%, higher than YOLOv11l trained on infrared images (60.52\%) but lower than the pure visible-light trained model (62.1\%). This indicates that multi-spectral images have key channels, and it's advisable to compare training results of both spectra before choosing the main branch.

Table \ref{tab:m3fd_fusion_strategies} shows the comparison of different fusion strategies on the M3FD dataset. For YOLOv11s, mid-fusion achieves the highest AP50 of 84.91\% and AP of 58.47\%, outperforming other strategies like early fusion (AP50 84.11\%, AP 58.19\%) and late fusion (AP50 84.63\%, AP 58.02\%). This aligns with previous studies of mid-level fusion strategies \cite{qingyun_cross-modality_2022,noauthor_adopting_nodate,tang_piafusion_2022, zhou_improving_2020,yan_cross-modality_2023}. However, for YOLOv11m, early fusion (AP50 87.06\%, AP 61.29\%) performs better than mid-fusion (AP50 86.71\%, AP 60.67\%). Moreover, the table reveals that most of the optimal detection results stem from early and mid-term fusion. This observation drove us to develop the P3-Midfusion method, as there might be a superior fusion strategy between early and mid-term fusion. Thus, while mid-fusion is often optimal, the best strategy can vary. Researchers and engineers should select fusion strategies based on their specific datasets and models. 

\begin{figure*}[htbp]
    \centering
    \includegraphics[width=0.8\textwidth]{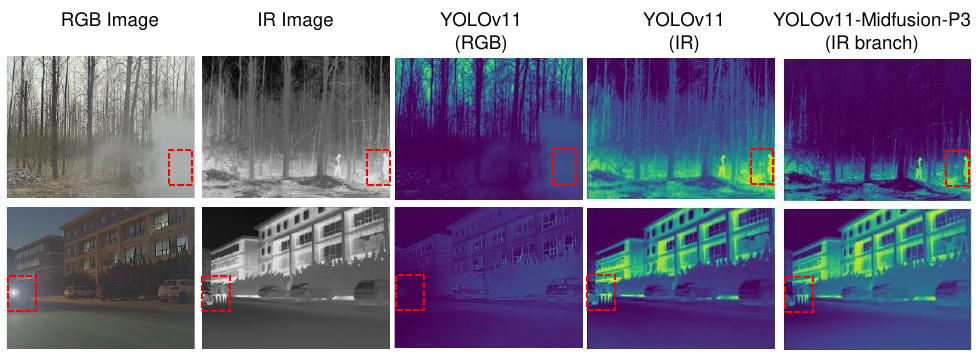}
    \caption{Feature maps visualization of multi-spectral fusion from stage2 (P2) of YOLOv11 model, illustrating enhanced object detection capabilities through combined rgb and infrared data processing.}
    \label{fig:fig7}
\end{figure*}

The feature map visualization in Figure \ref{fig:fig7} clearly shows the benefits of multi - spectral feature fusion. The feature maps shown are from stage2(P2) of the YOLOv11 model output, including RGB-only, IR-only, and mid-term fused RGB+IR feature maps. From the visualization, it's evident that models using only RGB or IR data can detect objects to a certain extent, but their detection capabilities are limited. For example, the RGB-only model may fail to recognize objects in low-visibility or smoky conditions. The IR - only model may miss objects that are not prominent in the infrared spectrum, leading to poorer detection performance than the pure RGB model, as shown in Table \ref{tab:m3fd_model_comparison_midfusion}. In contrast, the mid - term fusion model combining RGB and IR data demonstrates superior detection performance. Its feature maps not only highlight pedestrian outlines but also accurately show vehicles and other objects. This indicates that multi-spectral feature fusion can effectively integrate the advantages of different spectral bands, thereby significantly improving the model's detection accuracy and reliability.

% \subsection{Comparative experiments on KAIST and VEDAI datasets}
% TODO (maybe next Sunday... or next month, who knows? But for sure, not today!).

% The experiments are mostly done, but organizing the results takes time. The author, buried in projects, aims to release the second arXiv version this month.

\begin{figure*}[htbp]
    \centering
    \includegraphics[width=0.66\textwidth]{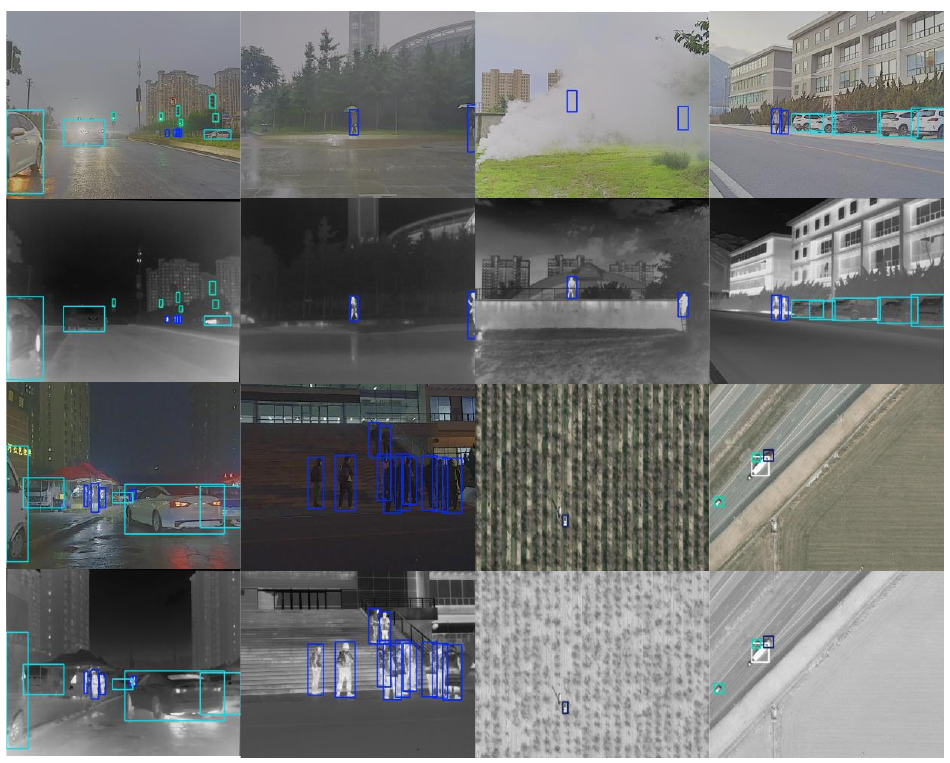}
    \caption{Some detection results on M3FD and VEDAI datasets of YOLOv11-RGBT-MCF.}
    \label{fig:fig8}
\end{figure*}

\subsection{Qualitative test}
\label{experiments:qualitative test}
We performed some qualitative results of YOLOv11-RGBT-MCF algorithm on two multispectral datasets, as given in Figure \ref{fig:fig8}. As depicted in the figure, the YOLOv11-RGBT-MCF model exhibites a strong capability in detecting objects in multispectral images, including those with complex backgrounds, low object discrimination, uneven lighting, smoke, rainy days, night time, as well as low-angle shooting perspectives, etc.

\section{Discussion}
The experiments in the above table prove the effectiveness, feasibility and generalisation of the model in the framework. In fact, in addition to the above experiments, we also designed a multispectral PGI \cite{wang_yolov9_2024} strategy and proposed several lightweight cross-attention mechanisms. Integrating it into YOLOv11-RGBT framework (see paper source address: \href{https://github.com/wandahangFY/YOLOv11-RGBT}{https://github.com/wandahangFY/YOLOv11-RGBT}). Multispectral PGI and cross-attention mechanisms can improve mAP by 0.5\% on some datasets, but we did not show it in the main trial because its improvement is limited and only effective on some datasets, which may stem from its dependence on specific spectral features. The distribution of spectral features in different datasets is different, which affects the utilization effect of PGI on gradient information. For example, the gradient guiding effect of PGI is more significant on datasets with distinct differences in spectral features. This suggests that whether to use these modules should be carefully chosen according to the specific data characteristics in practical applications. We also found that on some datasets, such as M3FD \cite{liu_target-aware_2022}, YOLOv11-midfusion model gets better detection results when the batch size is 32 than 16. For example, the mAP is about 0.6\% higher, but considering that all hyperparameters need to be consistent, Except for the batch size of model x, which is 8, all the remaining models are set to 16 as far as possible. Therefore, theoretically, there is still room for further improvement of some weights, and interested researchers can try in the future.

In addition, due to limited equipment resources, this paper only did the pre-training weight (from COCO \cite{lin_microsoft_2015} dataset)transfer training and multi-spectral controllable fine-tuning test of YOLOv11 on five datasets, and the other models only provide the experimental results without pre-training weights. Moreover, in order to ensure the generalization of the model, we did not introduce the attention mechanism  \cite{li_illumination-aware_2019, liu_multispectral_2016,qingyun_cross-modality_2022,sharma_yolors_2021,li_crossfuse_2024}
 and the low visibility module \cite{tang_divfusion_2023,tang_piafusion_2022,zhang_illumination-guided_2023} for experiments. In view of this, it is suggested that future research focus on improving the generalisation ability of the module and exploring adaptive adjustment strategies to adapt to multiple data sets and scenarios, so as to expand the scope of application of the module.

Despite some limitations, YOLOv11-RGBT framework has a wide application prospect in security monitoring, automatic driving and other fields with the advantages of multi-spectral fusion. Engineers can flexibly choose fusion modes and strategies according to specific scenario requirements. For future research, it is suggested to dig deeper into the intrinsic correlation of multi-spectral features and develop more efficient feature extraction and fusion methods. At the same time, the lightweight multi-spectral detection model is explored to reduce the hardware requirements, so as to promote the application of multi-spectral target detection technology in resource-constrained environments. We have open sourced most of the work mentioned in this paper, and will open source the weights and methods once the paper is published so that researchers and engineers can explore and improve it further.

\section{Conclusion}
Overall, We have developed YOLOv11-RGBT, a unified single-stage multimodal object detection framework. By re-evaluating the fusion strategy and the importance of the two modalities, and fully utilizing multispectral features, we've enhanced the model's generalization ability and detection performance. Experiments on three datasets confirm its effectiveness, providing new research ideas and methods for multispectral object detection and advancing the technology toward greater maturity and practicality. However, this paper hasn't deeply explored methods like multispectral PGI, lightweight cross-attention mechanisms, and low-light modules. Looking ahead, researchers can focus on improving module generalization and developing adaptive adjustment strategies. Exploring more efficient feature extraction, fusion methods, and lightweight model designs will expand the application scope of multispectral object detection. YOLOv11-RGBT has significant practical potential in security monitoring, autonomous driving, and other fields. Its efficiency and accuracy make it suitable for real-time applications like identifying threats in surveillance videos and detecting obstacles in autonomous vehicles.
 
Although our work has limitations in module performance and experimental scope due to equipment constraints, the framework shows great promise. In fact, our project supports not only multispectral data but also any image with pixel alignment, such as depth images, SAR images, etc. Moreover, we are conducting research on multispectral or multimodal tasks under non-aligned conditions and will make their specific implementations available to the public. Future work will continue to open-source more basic experiments, enhance the framework's capabilities, and explore new research directions, such as multispectral instance segmentation and keypoint detection. We also plan to apply our multispectral feature fusion concepts to other object detection algorithms and further study the practical deployment and application of our algorithm.

\section*{Acknowledgments}
This work was supported by the National Key Research and Development Program of China (No. 2023YFF0715502); Anhui Provincial Key Research and Development Project (No. 202304a05020013).

The authors thank the ultralytics team for the open source projects and also thank the editors and reviewers for their helpful suggestions.

%Bibliography
\bibliographystyle{unsrt}  
\bibliography{YOLOv11-RGBT-arxiv}

\begin{thebibliography}{10}

\bibitem{wan_yolo-mif_2024}
Dahang Wan, Rongsheng Lu, Bingtao Hu, Jiajie Yin, Siyuan Shen, Ting xu, and Xianli Lang.
\newblock {YOLO}-{MIF}: Improved {YOLOv}8 with multi-information fusion for object detection in gray-scale images.
\newblock 62:102709.

\bibitem{liu_coconet_2024}
Jinyuan Liu, Runjia Lin, Guanyao Wu, Risheng Liu, Zhongxuan Luo, and Xin Fan.
\newblock {CoCoNet}: Coupled contrastive learning network with multi-level feature ensemble for multi-modality image fusion.
\newblock 132(5):1748--1775.

\bibitem{liu_multi-interactive_2023}
Jinyuan Liu, Zhu Liu, Guanyao Wu, Long Ma, Risheng Liu, Wei Zhong, Zhongxuan Luo, and Xin Fan.
\newblock Multi-interactive feature learning and a full-time multi-modality benchmark for image fusion and segmentation.
\newblock In {\em 2023 {IEEE}/{CVF} International Conference on Computer Vision ({ICCV})}, pages 8081--8090.
\newblock {ISSN}: 2380-7504.

\bibitem{liu_target-aware_2022}
Jinyuan Liu, Xin Fan, Zhanbo Huang, Guanyao Wu, Risheng Liu, Wei Zhong, and Zhongxuan Luo.
\newblock Target-aware dual adversarial learning and a multi-scenario multi-modality benchmark to fuse infrared and visible for object detection.
\newblock In {\em 2022 {IEEE}/{CVF} Conference on Computer Vision and Pattern Recognition ({CVPR})}, pages 5792--5801.
\newblock {ISSN}: 2575-7075.

\bibitem{hwang_multispectral_2015}
Soonmin Hwang, Jaesik Park, Namil Kim, Yukyung Choi, and In~So Kweon.
\newblock Multispectral pedestrian detection: Benchmark dataset and baseline.
\newblock In {\em 2015 {IEEE} Conference on Computer Vision and Pattern Recognition ({CVPR})}, pages 1037--1045. {IEEE}.

\bibitem{redmon_you_2016}
Joseph Redmon, Santosh Divvala, Ross Girshick, and Ali Farhadi.
\newblock You only look once: Unified, real-time object detection.

\bibitem{redmon_yolo9000_2016}
Joseph Redmon and Ali Farhadi.
\newblock {YOLO}9000: Better, faster, stronger.

\bibitem{redmon_yolov3_2018}
Joseph Redmon and Ali Farhadi.
\newblock {YOLOv}3: An incremental improvement.

\bibitem{bochkovskiy_yolov4_2020}
Alexey Bochkovskiy, Chien-Yao Wang, and Hong-Yuan~Mark Liao.
\newblock {YOLOv}4: Optimal speed and accuracy of object detection.

\bibitem{noauthor_ultralyticsyolov5_2022}
ultralytics/yolov5.
\newblock original-date: 2020-05-18T03:45:11Z.

\bibitem{li_yolov6_2022}
Chuyi Li, Lulu Li, Hongliang Jiang, Kaiheng Weng, Yifei Geng, Liang Li, Zaidan Ke, Qingyuan Li, Meng Cheng, Weiqiang Nie, Yiduo Li, Bo~Zhang, Yufei Liang, Linyuan Zhou, Xiaoming Xu, Xiangxiang Chu, Xiaoming Wei, and Xiaolin Wei.
\newblock {YOLOv}6: A single-stage object detection framework for industrial applications.

\bibitem{wang_yolov7_2023}
Chien-Yao Wang, Alexey Bochkovskiy, and Hong-Yuan~Mark Liao.
\newblock {YOLOv}7: Trainable bag-of-freebies sets new state-of-the-art for real-time object detectors.
\newblock pages 7464--7475.

\bibitem{jocher_yolo_2023}
Glenn Jocher, Ayush Chaurasia, and Jing Qiu.
\newblock {YOLO} by ultralytics.
\newblock original-date: 2022-09-11T16:39:45Z.

\bibitem{wang_yolov9_2024}
Chien-Yao Wang, I.-Hau Yeh, and Hong-Yuan~Mark Liao.
\newblock {YOLOv}9: Learning what you want to learn using programmable gradient information.
\newblock version: 1.

\bibitem{wang_yolov10_2024}
Ao~Wang, Hui Chen, Lihao Liu, Kai Chen, Zijia Lin, Jungong Han, and Guiguang Ding.
\newblock {YOLOv}10: Real-time end-to-end object detection.

\bibitem{khanam_yolov11_2024}
Rahima Khanam and Muhammad Hussain.
\newblock {YOLOv}11: An overview of the key architectural enhancements.

\bibitem{tian_yolov12_2025}
Yunjie Tian, Qixiang Ye, and David Doermann.
\newblock {YOLOv}12: Attention-centric real-time object detectors.

\bibitem{leibe_ssd_2016}
Wei Liu, Dragomir Anguelov, Dumitru Erhan, Christian Szegedy, Scott Reed, Cheng-Yang Fu, and Alexander~C. Berg.
\newblock {SSD}: Single shot {MultiBox} detector.
\newblock In Bastian Leibe, Jiri Matas, Nicu Sebe, and Max Welling, editors, {\em Computer Vision – {ECCV} 2016}, volume 9905, pages 21--37. Springer International Publishing.
\newblock Series Title: Lecture Notes in Computer Science.

\bibitem{fu_dssd_nodate}
Cheng-Yang Fu, Wei Liu, Ananth Ranga, Ambrish Tyagi, and Alexander~C Berg.
\newblock {DSSD} : Deconvolutional single shot detector.
\newblock page~11.

\bibitem{girshick_rich_2014}
Ross Girshick, Jeff Donahue, Trevor Darrell, and Jitendra Malik.
\newblock Rich feature hierarchies for accurate object detection and semantic segmentation.
\newblock version: 5.

\bibitem{girshick_fast_2015}
Ross Girshick.
\newblock Fast r-{CNN}.

\bibitem{ren_faster_2017}
Shaoqing Ren, Kaiming He, Ross Girshick, and Jian Sun.
\newblock Faster r-{CNN}: Towards real-time object detection with region proposal networks.
\newblock 39(6):1137--1149.
\newblock Conference Name: {IEEE} Transactions on Pattern Analysis and Machine Intelligence.

\bibitem{cai_cascade_2018}
Zhaowei Cai and Nuno Vasconcelos.
\newblock Cascade r-{CNN}: Delving into high quality object detection.
\newblock In {\em 2018 {IEEE}/{CVF} Conference on Computer Vision and Pattern Recognition}, pages 6154--6162.
\newblock {ISSN}: 2575-7075.

\bibitem{li_pixel-level_2017}
Shutao Li, Xudong Kang, Leyuan Fang, Jianwen Hu, and Haitao Yin.
\newblock Pixel-level image fusion: A survey of the state of the art.
\newblock 33:100--112.

\bibitem{dong_mdcnn_2021}
Meilin Dong, Weisheng Li, Xuesong Liang, and Xiayan Zhang.
\newblock {MDCNN}: multispectral pansharpening based on a multiscale dilated convolutional neural network.
\newblock 15(3):036516.
\newblock Publisher: {SPIE}.

\bibitem{li_crossfuse_2024}
Hui Li and Xiao-Jun Wu.
\newblock {CrossFuse}: A novel cross attention mechanism based infrared and visible image fusion approach.
\newblock 103:102147.

\bibitem{tang_divfusion_2023}
Linfeng Tang, Xinyu Xiang, Hao Zhang, Meiqi Gong, and Jiayi Ma.
\newblock {DIVFusion}: Darkness-free infrared and visible image fusion.
\newblock 91:477--493.

\bibitem{li_illumination-aware_2019}
Chengyang Li, Dan Song, Ruofeng Tong, and Min Tang.
\newblock Illumination-aware faster r-{CNN} for robust multispectral pedestrian detection.
\newblock 85:161--171.

\bibitem{liu_multispectral_2016}
Jingjing Liu, Shaoting Zhang, Shu Wang, and Dimitris~N. Metaxas.
\newblock Multispectral deep neural networks for pedestrian detection.

\bibitem{sharma_yolors_2021}
Manish Sharma, Mayur Dhanaraj, Srivallabha Karnam, Dimitris~G. Chachlakis, Raymond Ptucha, Panos~P. Markopoulos, and Eli Saber.
\newblock {YOLOrs}: Object detection in multimodal remote sensing imagery.
\newblock 14:1497--1508.
\newblock Conference Name: {IEEE} Journal of Selected Topics in Applied Earth Observations and Remote Sensing.

\bibitem{qingyun_cross-modality_2022}
Fang Qingyun, Han Dapeng, and Wang Zhaokui.
\newblock Cross-modality fusion transformer for multispectral object detection.

\bibitem{fang_cross-modality_2021}
Qingyun Fang and Zhaokui Wang.
\newblock Cross-modality attentive feature fusion for object detection in multispectral remote sensing imagery.

\bibitem{shen_icafusion_2024}
Jifeng Shen, Yifei Chen, Yue Liu, Xin Zuo, Heng Fan, and Wankou Yang.
\newblock {ICAFusion}: Iterative cross-attention guided feature fusion for multispectral object detection.
\newblock 145:109913.

\bibitem{li_multiscale_2024}
Ruimin Li, Jiajun Xiang, Feixiang Sun, Ye~Yuan, Longwu Yuan, and Shuiping Gou.
\newblock Multiscale cross-modal homogeneity enhancement and confidence-aware fusion for multispectral pedestrian detection.
\newblock 26:852--863.
\newblock Conference Name: {IEEE} Transactions on Multimedia.

\bibitem{zhou_improving_2020}
Kailai Zhou, Linsen Chen, and Xun Cao.
\newblock Improving multispectral pedestrian detection by addressing modality imbalance problems.

\bibitem{tang_piafusion_2022}
Linfeng Tang, Jiteng Yuan, Hao Zhang, Xingyu Jiang, and Jiayi Ma.
\newblock {PIAFusion}: A progressive infrared and visible image fusion network based on illumination aware.
\newblock 83-84:79--92.

\bibitem{zhang_illumination-guided_2023}
Yan Zhang, Huai Yu, Yujie He, Xinya Wang, and Wen Yang.
\newblock Illumination-guided {RGBT} object detection with inter- and intra-modality fusion.
\newblock 72:1--13.
\newblock Conference Name: {IEEE} Transactions on Instrumentation and Measurement.

\bibitem{woo_cbam_2018}
Sanghyun Woo, Jongchan Park, Joon-Young Lee, and In~So Kweon.
\newblock {CBAM}: Convolutional block attention module.

\bibitem{wan_mixed_2023}
Dahang Wan, Rongsheng Lu, Siyuan Shen, Ting Xu, Xianli Lang, and Zhijie Ren.
\newblock Mixed local channel attention for object detection.
\newblock 123:106442.

\bibitem{zhang_multispectral_2020}
Heng Zhang, Elisa Fromont, Sébastien Lefevre, and Bruno Avignon.
\newblock Multispectral fusion for object detection with cyclic fuse-and-refine blocks.

\bibitem{razakarivony_vehicle_2016}
Sebastien Razakarivony and Frederic Jurie.
\newblock Vehicle detection in aerial imagery : A small target detection benchmark.
\newblock 34:187--203.

\bibitem{noauthor_adopting_nodate}
Adopting the {YOLOv}4 architecture for low-latency multispectral pedestrian detection in autonomous driving.

\bibitem{yan_cross-modality_2023}
Chaoqi Yan, Hong Zhang, Xuliang Li, Yifan Yang, and Ding Yuan.
\newblock Cross-modality complementary information fusion for multispectral pedestrian detection.
\newblock 35(14):10361--10386.

\bibitem{jia_llvip_2021}
Xinyu Jia, Chuang Zhu, Minzhen Li, Wenqi Tang, and Wenli Zhou.
\newblock {LLVIP}: A visible-infrared paired dataset for low-light vision.
\newblock In {\em 2021 {IEEE}/{CVF} International Conference on Computer Vision Workshops ({ICCVW})}, pages 3489--3497.
\newblock {ISSN}: 2473-9944.

\bibitem{choi_kaist_2018}
Yukyung Choi, Namil Kim, Soonmin Hwang, Kibaek Park, Jae~Shin Yoon, Kyounghwan An, and In~So Kweon.
\newblock {KAIST} multi-spectral day/night data set for autonomous and assisted driving.
\newblock 19(3):934--948.

\bibitem{lin_focal_2018}
Tsung-Yi Lin, Priya Goyal, Ross Girshick, Kaiming He, and Piotr Dollár.
\newblock Focal loss for dense object detection.

\bibitem{song_sun_2015}
Shuran Song, Samuel~P. Lichtenberg, and Jianxiong Xiao.
\newblock {SUN} {RGB}-d: A {RGB}-d scene understanding benchmark suite.
\newblock pages 567--576.

\bibitem{toker_dynamicearthnet_nodate}
Aysim Toker, Lukas Kondmann, Mark Weber, Marvin Eisenberger, Andres Camero, Jingliang Hu, Ariadna~Pregel Hoderlein, Caglar Senaras, Timothy Davis, Daniel Cremers, Giovanni Marchisio, Xiao~Xiang Zhu, and Laura Leal-Taixe.
\newblock {DynamicEarthNet}: Daily multi-spectral satellite dataset for semantic change segmentation.

\bibitem{guo_damsdet_nodate}
Junjie Guo, Chenqiang Gao, Fangcen Liu, Deyu Meng, and Xinbo Gao.
\newblock {DAMSDet}: Dynamic adaptive multispectral detection transformer with competitive query selection and adaptive feature fusion.

\bibitem{sun_investigating_2019}
Weiwei Sun, Kai Ren, Gang Yang, Xiangchao Meng, and Yinnian Liu.
\newblock Investigating {GF}-5 hyperspectral and {GF}-1 multispectral data fusion methods for multitemporal change analysis.
\newblock In {\em 2019 10th International Workshop on the Analysis of Multitemporal Remote Sensing Images ({MultiTemp})}, pages 1--4.

\bibitem{li_image_2013}
Shutao Li, Xudong Kang, and Jianwen Hu.
\newblock Image fusion with guided filtering.
\newblock 22(7):2864--2875.
\newblock Conference Name: {IEEE} Transactions on Image Processing.

\bibitem{li_detail_2021}
Ling Li, Yaochen Li, Chuan Wu, Hang Dong, Peilin Jiang, and Fei Wang.
\newblock Detail fusion {GAN}: High-quality translation for unpaired images with {GAN}-based data augmentation.
\newblock In {\em 2020 25th International Conference on Pattern Recognition ({ICPR})}, pages 1731--1736.
\newblock {ISSN}: 1051-4651.

\bibitem{xue_maf-yolo_2021}
Yongjie Xue, Zhiyong Ju, Yuming Li, and Wenxin Zhang.
\newblock {MAF}-{YOLO}: Multi-modal attention fusion based {YOLO} for pedestrian detection.
\newblock 118:103906.

\bibitem{zhang_learning_2018}
Xingming Zhang, Xuehan Zhang, Xuedan Du, Xiangming Zhou, and Jun Yin.
\newblock Learning multi-domain convolutional network for {RGB}-t visual tracking.
\newblock In {\em 2018 11th International Congress on Image and Signal Processing, {BioMedical} Engineering and Informatics ({CISP}-{BMEI})}, pages 1--6.

\bibitem{zhao_detrs_2024}
Yian Zhao, Wenyu Lv, Shangliang Xu, Jinman Wei, Guanzhong Wang, Qingqing Dang, Yi~Liu, and Jie Chen.
\newblock {DETRs} beat {YOLOs} on real-time object detection.
\newblock In {\em 2024 {IEEE}/{CVF} Conference on Computer Vision and Pattern Recognition ({CVPR})}, pages 16965--16974.
\newblock {ISSN}: 2575-7075.

\bibitem{xu_pp-yoloe_2022}
Shangliang Xu, Xinxin Wang, Wenyu Lv, Qinyao Chang, Cheng Cui, Kaipeng Deng, Guanzhong Wang, Qingqing Dang, Shengyu Wei, Yuning Du, and Baohua Lai.
\newblock {PP}-{YOLOE}: An evolved version of {YOLO}.

\bibitem{garzelli_multispectral_2018}
Andrea Garzelli, Bruno Aiazzi, Luciano Alparone, Simone Lolli, and Gemine Vivone.
\newblock Multispectral pansharpening with radiative transfer-based detail-injection modeling for preserving changes in vegetation cover.
\newblock 10(8):1308.
\newblock Number: 8 Publisher: Multidisciplinary Digital Publishing Institute.

\bibitem{zhang_rethinking_2024}
Xue Zhang, Si-Yuan Cao, Fang Wang, Runmin Zhang, Zhe Wu, Xiaohan Zhang, Xiaokai Bai, and Hui-Liang Shen.
\newblock Rethinking early-fusion strategies for improved multispectral object detection.
\newblock pages 1--15.

\bibitem{zhang_adding_2023}
Lvmin Zhang, Anyi Rao, and Maneesh Agrawala.
\newblock Adding conditional control to text-to-image diffusion models.
\newblock pages 3836--3847.

\bibitem{lin_microsoft_2015}
Tsung-Yi Lin, Michael Maire, Serge Belongie, Lubomir Bourdev, Ross Girshick, James Hays, Pietro Perona, Deva Ramanan, C.~Lawrence Zitnick, and Piotr Dollár.
\newblock Microsoft {COCO}: Common objects in context.

\bibitem{zhang_tfdet_2024}
Xue Zhang, Xiaohan Zhang, Jiangtao Wang, Jiacheng Ying, Zehua Sheng, Heng Yu, Chunguang Li, and Hui-Liang Shen.
\newblock {TFDet}: Target-aware fusion for {RGB}-t pedestrian detection.
\newblock pages 1--15.

\bibitem{chen_multimodal_2022}
Yi-Ting Chen, Jinghao Shi, Zelin Ye, Christoph Mertz, Deva Ramanan, and Shu Kong.
\newblock Multimodal object detection via probabilistic ensembling.
\newblock In Shai Avidan, Gabriel Brostow, Moustapha Cissé, Giovanni~Maria Farinella, and Tal Hassner, editors, {\em Computer Vision – {ECCV} 2022}, pages 139--158. Springer Nature Switzerland.

\bibitem{you_multi-scale_2023}
Shuai You, Xuedong Xie, Yujian Feng, Chaojun Mei, and Yimu Ji.
\newblock Multi-scale aggregation transformers for multispectral object detection.
\newblock 30:1172--1176.

\bibitem{cao_multi-modal_2023}
Bing Cao, Yiming Sun, Pengfei Zhu, and Qinghua Hu.
\newblock Multi-modal gated mixture of local-to-global experts for dynamic image fusion.
\newblock In {\em 2023 {IEEE}/{CVF} International Conference on Computer Vision ({ICCV})}, pages 23498--23507. {IEEE}.

\bibitem{cao_multimodal_2023}
Yue Cao, Junchi Bin, Jozsef Hamari, Erik Blasch, and Zheng Liu.
\newblock Multimodal object detection by channel switching and spatial attention.
\newblock In {\em 2023 {IEEE}/{CVF} Conference on Computer Vision and Pattern Recognition Workshops ({CVPRW})}, pages 403--411.
\newblock {ISSN}: 2160-7516.

\bibitem{zhu_multi-modal_2023}
Yaohui Zhu, Xiaoyu Sun, Miao Wang, and Hua Huang.
\newblock Multi-modal feature pyramid transformer for {RGB}-infrared object detection.
\newblock 24(9):9984--9995.

\bibitem{zhang_cmx_2023}
Jiaming Zhang, Huayao Liu, Kailun Yang, Xinxin Hu, Ruiping Liu, and Rainer Stiefelhagen.
\newblock {CMX}: Cross-modal fusion for {RGB}-x semantic segmentation with transformers.
\newblock 24(12):14679--14694.

\bibitem{fu_lraf-net_2024}
Haolong Fu, Shixun Wang, Puhong Duan, Changyan Xiao, Renwei Dian, Shutao Li, and Zhiyong Li.
\newblock {LRAF}-net: Long-range attention fusion network for visible–infrared object detection.
\newblock 35(10):13232--13245.
\newblock Conference Name: {IEEE} Transactions on Neural Networks and Learning Systems.

\bibitem{chen_igt_2023}
Keyu Chen, Jinqiang Liu, and Han Zhang.
\newblock {IGT}: Illumination-guided {RGB}-t object detection with transformers.
\newblock 268:110423.

\bibitem{fu_yolo-adaptor_2024}
Haolong Fu, Hanhao Liu, Jin Yuan, Xuan He, Jiacheng Lin, and Zhiyong Li.
\newblock {YOLO}-adaptor: A fast adaptive one-stage detector for non-aligned visible-infrared object detection.
\newblock pages 1--14.

\bibitem{dong_fusion-mamba_2024}
Wenhao Dong, Haodong Zhu, Shaohui Lin, Xiaoyan Luo, Yunhang Shen, Xuhui Liu, Juan Zhang, Guodong Guo, and Baochang Zhang.
\newblock Fusion-mamba for cross-modality object detection.

\bibitem{ma_fusiongan_2019}
Jiayi Ma, Wei Yu, Pengwei Liang, Chang Li, and Junjun Jiang.
\newblock {FusionGAN}: A generative adversarial network for infrared and visible image fusion.
\newblock 48:11--26.

\bibitem{li_densefuse_2019}
Hui Li and Xiao-Jun Wu.
\newblock {DenseFuse}: A fusion approach to infrared and visible images.
\newblock 28(5):2614--2623.

\bibitem{xu_u2fusion_2022}
Han Xu, Jiayi Ma, Junjun Jiang, Xiaojie Guo, and Haibin Ling.
\newblock U2fusion: A unified unsupervised image fusion network.
\newblock 44(1):502--518.

\bibitem{zhang_weakly_2025}
Lu~Zhang, Zhiyong Liu, Xiangyu Zhu, Zhan Song, Xu~Yang, Zhen Lei, and Hong Qiao.
\newblock Weakly aligned feature fusion for multimodal object detection.
\newblock 36(3):4145--4159.

\bibitem{zhao_ddfm_2023}
Zixiang Zhao, Haowen Bai, Yuanzhi Zhu, Jiangshe Zhang, Shuang Xu, Yulun Zhang, Kai Zhang, Deyu Meng, Radu Timofte, and Luc Van~Gool.
\newblock {DDFM}: Denoising diffusion model for multi-modality image fusion.
\newblock In {\em 2023 {IEEE}/{CVF} International Conference on Computer Vision ({ICCV})}, pages 8048--8059. {IEEE}.

\bibitem{xie_learning_2022}
Jin Xie, Rao~Muhammad Anwer, Hisham Cholakkal, Jing Nie, Jiale Cao, Jorma Laaksonen, and Fahad~Shahbaz Khan.
\newblock Learning a dynamic cross-modal network for multispectral pedestrian detection.
\newblock In {\em Proceedings of the 30th {ACM} International Conference on Multimedia}, {MM} '22, pages 4043--4052. Association for Computing Machinery.

\bibitem{sun_detfusion_2022}
Yiming Sun, Bing Cao, Pengfei Zhu, and Qinghua Hu.
\newblock {DetFusion}: A detection-driven infrared and visible image fusion network.
\newblock In {\em Proceedings of the 30th {ACM} International Conference on Multimedia}, pages 4003--4011. {ACM}.

\bibitem{cao_lightweight_2023}
Yue Cao, Yanshuo Fan, Junchi Bin, and Zheng Liu.
\newblock Lightweight transformer for multi-modal object detection (student abstract).
\newblock In {\em Proceedings of the Thirty-Seventh {AAAI} Conference on Artificial Intelligence and Thirty-Fifth Conference on Innovative Applications of Artificial Intelligence and Thirteenth Symposium on Educational Advances in Artificial Intelligence}. {AAAI} Press.

\bibitem{tang_divfusion_2023-1}
Linfeng Tang, Xinyu Xiang, Hao Zhang, Meiqi Gong, and Jiayi Ma.
\newblock {DIVFusion}: Darkness-free infrared and visible image fusion.
\newblock 91:477--493.

\bibitem{xu_dm-fusion_2024}
Guoxia Xu, Chunming He, Hao Wang, Hu~Zhu, and Weiping Ding.
\newblock {DM}-fusion: Deep model-driven network for heterogeneous image fusion.
\newblock 35(7):10071--10085.

\bibitem{he_multispectral_2023}
Xiao He, Chang Tang, Xin Zou, and Wei Zhang.
\newblock Multispectral object detection via cross-modal conflict-aware learning.
\newblock In {\em Proceedings of the 31st {ACM} International Conference on Multimedia}, {MM} '23, pages 1465--1474. Association for Computing Machinery.

\bibitem{zhao_metafusion_2023}
Wenda Zhao, Shigeng Xie, Fan Zhao, You He, and Huchuan Lu.
\newblock {MetaFusion}: Infrared and visible image fusion via meta-feature embedding from object detection.
\newblock In {\em 2023 {IEEE}/{CVF} Conference on Computer Vision and Pattern Recognition ({CVPR})}, pages 13955--13965.
\newblock {ISSN}: 2575-7075.

\bibitem{chu_toward_2024}
Fuchen Chu, Jiale Cao, Zhanjie Song, Zhuang Shao, Yanwei Pang, and Xuelong Li.
\newblock Toward generalizable multispectral pedestrian detection.
\newblock 25(5):3739--3750.

\bibitem{tang_camf_2024}
Linfeng Tang, Ziang Chen, Jun Huang, and Jiayi Ma.
\newblock {CAMF}: An interpretable infrared and visible image fusion network based on class activation mapping.
\newblock 26:4776--4791.

\bibitem{chen_lenfusion_2024}
Jun Chen, Liling Yang, Wei Liu, Xin Tian, and Jiayi Ma.
\newblock {LENFusion}: A joint low-light enhancement and fusion network for nighttime infrared and visible image fusion.
\newblock 73:1--15.

\bibitem{yi_diff-if_2024}
Xunpeng Yi, Linfeng Tang, Hao Zhang, Han Xu, and Jiayi Ma.
\newblock Diff-{IF}: Multi-modality image fusion via diffusion model with fusion knowledge prior.
\newblock 110:102450.

\end{thebibliography}

\end{document}